\documentclass[runningheads]{llncs}

\usepackage{eccv}

\usepackage{eccvabbrv}

\usepackage{graphicx}
\usepackage{booktabs}
\usepackage{sidecap}

\usepackage[table]{xcolor}
\usepackage{multicol}
\usepackage{multirow}
\usepackage{mathrsfs}

\usepackage[accsupp]{axessibility}  

\usepackage[ruled,vlined]{algorithm2e}

\newcommand\blfootnote[1]{%
  \begingroup
  \renewcommand\thefootnote{}\footnotetext{#1}%
  \addtocounter{footnote}{-1}%
  \endgroup
}

\usepackage{hyperref}

\usepackage{orcidlink}

\begin{document}

\title{Ex-Sim(3)-Reg: 2D-3D Correspondence Pruning via Extended Sim(3) Registration} 

\titlerunning{2D-3D Correspondence Pruning via Extended Sim(3) Registration}

\author{Pei An\inst{1*}\orcidlink{0000-0002-3645-8465} \and
Muyao Peng \inst{1*}\orcidlink{0009-0004-9825-5307}
\and
Junfeng Ding \inst{1}\orcidlink{0000-0001-7328-9923}
\and
Jiaqi Yang \inst{2\dag}\orcidlink{0000-0002-2071-2457}
\and\\
Liangliang Nan \inst{3\dag}\orcidlink{0000-0002-5629-9975}
}

\authorrunning{P.~An et al.}

\institute{
Huazhong University of Science and Technology, China 
\email{\{anpei96,muyao99,djfenghust\}@hust.edu.cn}
\and
Northwestern Polytechnical University, China 
\email{jqyang@nwpu.edu.cn}
\and
Delft University of Technology, Netherlands 
\email{liangliang.nan@tudelft.nl}
}
\maketitle

\blfootnote{
$*$ Equal contribution. \\
$\dag$ Corresponding author(s).
}

\begin{abstract}


Learning-based image-to-point-cloud (I2P) registration has garnered increasing attention in recent years. Nevertheless, existing methods still struggle with severe outliers under challenging scenarios with unseen, low-inlier, or distorted cases. A fast and robust 2D-3D correspondence pruning method is therefore highly desirable. Recently, a promising scheme lifts 2D-3D correspondences to 3D-3D correspondences using depth priors, casting correspondence pruning as a Sim(3) registration problem. However, depth priors estimated from monocular images are inherently noisy, which undermines the reliability of this scheme. In this paper, to explicitly model non-negligible depth noise, we reformulate correspondence pruning as an extended Sim(3) registration problem and propose a simple yet effective pruning algorithm termed Ex-Sim(3)-Reg. We further provide a theoretical analysis to justify the effectiveness of our method. Extensive experiments on the 7-Scenes, RGBD-V2, ScanNet, and TUM datasets demonstrate that Ex-Sim(3)-Reg achieves up to \textbf{24.7\% improvement} in registration recall over state-of-the-art baseline methods. Code is released at \url{github.com/anpei96/ex-sim3-demo}.

\end{abstract}


\section{Introduction}
\label{sec.intro}

Image-to-point-cloud (I2P) registration \cite{2d3d-matr} enables a wide range of applications in 3D computer vision, such as visual localization, 3D reconstruction, and multi-sensor calibration~\cite{cmr-next, mast3r-slam, 2d3d-match, CoFiI2P}. Despite rapid advances in deep learning-based I2P registration~\cite{2d3d-matr, bridge, corr-learning-i2p-cvpr}, current methods still struggle to generalize to unseen, diverse, and complex real-world scenarios~\cite{top-i2p, new_add_1}. Specifically, they often predict a large number of incorrect 2D-3D correspondences (i.e., outliers), resulting in instability of downstream tasks~\cite{graph-i2p-cvpr, softposit}.

To mitigate outliers, it is crucial to study the 2D-3D correspondence pruning problem. RANSAC-based P3P \cite{ransac-p3p} serves as a default pruning algorithm in most learning-based I2P pipelines \cite{2d3d-matr, bridge}. However, when predicted correspondences are heavily contaminated by outliers, RANSAC-based P3P frequently fails to compute an accurate camera pose. Some works formulate 2D-3D correspondence pruning as a non-convex optimization problem~\cite{softposit, diff-pnp-layer, blind-pnp, bnb-blind-pnp-1}. Yet these methods are either computationally expensive~\cite{softposit, bnb-blind-pnp-1} or require additional model learning~\cite{diff-pnp-layer, blind-pnp}, thereby limiting their practicality. Recently, a more practical strategy has emerged that lifts 2D-3D correspondences to 3D-3D correspondences with depth priors~\cite{graph-i2p-cvpr, mast3r-slam, depth-shift-2, freereg}. This approach offers two advantages: (i) monocular depth priors can be readily predicted with existing visual foundation models \cite{dep_anything_v2, mast3r}; (ii) metric depth formulates correspondence pruning as a Sim(3) registration problem, which can be effectively solved using classical geometric methods {without deep learning} \cite{depth-shift-2}. 


Although depth prior-based pruning is attractive, existing methods rely heavily on highly accurate depth priors \cite{mast3r-slam, freereg, depth-shift-2, graph-i2p-cvpr}. Due to the inherent scale ambiguity of monocular images, depth priors inevitably introduce \textbf{non-negligible} errors \cite{depth-shift-1, depth-shift-3}, leading to degraded I2P registration results.


\begin{figure}[t]
	\centering
		\includegraphics[width=1.0\linewidth]{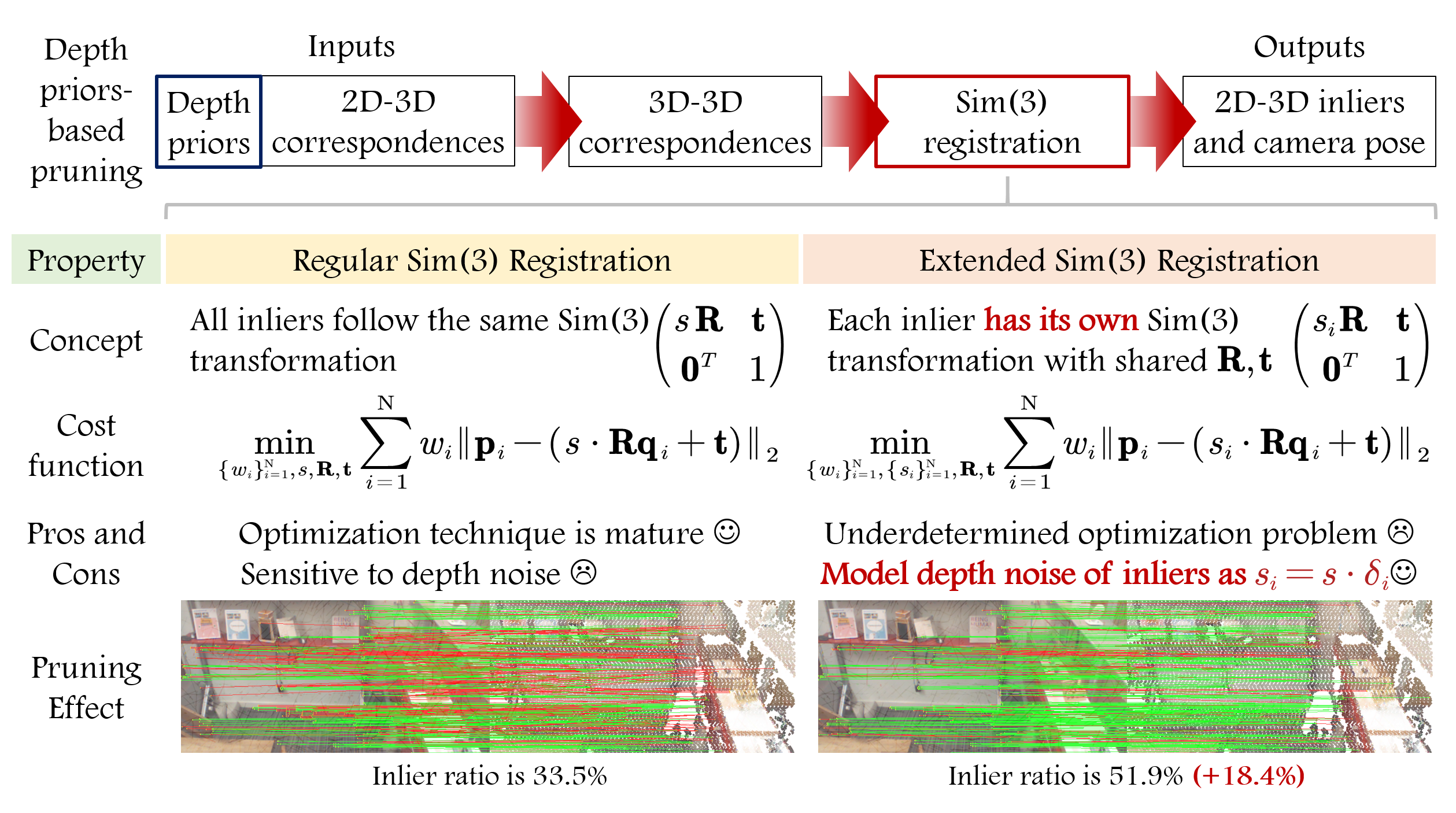}	
		\caption{Motivation. We extend the Sim(3) registration to model the noise $\delta_i$ in depth priors as correspondence-specific scale perturbations $s_i=s \cdot \delta_i$. This simple reformulation significantly improves pruning robustness against inaccurate depth priors.}
	\label{fig:intro}
    \vspace{-12pt}
\end{figure}


To address this issue, by modeling the depth noise as correspondence-specific scale perturbations (Fig. \ref{fig:intro}), we reformulate pruning from a regular Sim(3) to an \textbf{extended Sim(3)} registration problem. It compensates for depth noise during inlier optimization, which theoretically improves the pruning robustness to the inaccurate depth priors. To solve this underdetermined problem, we develop a heuristic yet effective pruning algorithm named Ex-Sim(3)-Reg. Inliers are identified by searching for maximum cliques in a family of extended compatibility graphs parameterized by scale hypotheses. Moreover, we theoretically explain the effectiveness of Ex-Sim(3)-Reg by analyzing the structure of extended compatibility graphs. In practice, Ex-Sim(3)-Reg can combine with RANSAC-based P3P \cite{ransac-p3p} for powerful 2D-3D pruning. Results on the 7-Scenes \cite{scene-7-dataset}, RGBD-V2 \cite{RGBD-dataset}, ScanNet \cite{scan-net}, and TUM \cite{tum} datasets demonstrate up to 24.7\% gain in registration recall over baseline pruning methods. 

Our main contributions are summarized as follows:


\begin{itemize}
    \item We are \textbf{the first} to reformulate depth prior-based 2D-3D correspondence pruning as an extended Sim(3) registration problem that models depth noise as correspondence-specific scale perturbations.
    \item We propose Ex-Sim(3)-Reg, a fast, non-learning pruning algorithm that approximates the extended Sim(3) registration problem via scale-parameterized compatibility graphs, together with a theoretical analysis of its effectiveness.
    \item In practice, Ex-Sim(3)-Reg can work with RANSAC-based P3P to achieve better performance than state-of-the-art methods on multiple datasets. 
\end{itemize}

\section{Related Works}
\label{sec.works}

Up to now, I2P registration typically comprises two procedures: (i) I2P feature matching to generate candidate 2D-3D correspondences, and (ii) 2D-3D correspondence pruning to remove outliers during pose estimation.

\noindent \textbf{I2P feature matching.} Current methods for generating I2P correspondences fall into two categories: descriptor-based and flow-based schemes~\cite{cmr-next}. Descriptor-based methods learn cross-modal features for 2D pixels and 3D points and match them via nearest neighbor search in the feature space~\cite{2d3d-matr, 2d3d-match, p2-net, top-i2p}. For example, P2-Net~\cite{p2-net} uses separate backbones to extract image and point cloud features supervised by a circle loss~\cite{circle-loss} to improve the correspondence accuracy. Follow-up works improve stability by leveraging patch-level correspondences~\cite{2d3d-matr} or by incorporating topological cues~\cite{top-i2p}. Bie et al.~\cite{graph-i2p-cvpr} reduces the cross-modal gap by incorporating metric depth predicted from monocular images. Flow-based methods predict dense pixel-wise displacements between an image and a projected point cloud map and are particularly prevalent for outdoor or LiDAR-camera setups~\cite{cmr-agent, cmr-next, cmrnet, i2d-loc}. Recent work (i.e., CMRNext~\cite{cmr-next}, 2025) unifies descriptor and flow ideas to boost localization in prior LiDAR maps. In summary, the framework of I2P feature matching has been fully developed in recent years.

\noindent \textbf{2D-3D correspondence pruning.} Pruning removes mismatches produced by the matching stage and is critical for robust I2P registration. Early approaches employ global optimization (e.g., branch-and-bound \cite{bnb-blind-pnp-1}), but these methods are computationally prohibitive for many applications~\cite{blind-pnp}. After 2020, learning-based approaches have been proposed, but most require high initial inlier ratios or additional viewpoint priors~\cite{blind-pnp,gomatch}. Recently, depth prior-based pruning gained increasing attention, because it allows formulating the original problem as a Sim(3) registration. Bie et al.~\cite{graph-i2p-cvpr}  developed a graph neural network-based method to identify inliers from the Sim(3) registration problem. Murai et al.~\cite{mast3r-slam} solved the Sim(3) registration problem by the iterative closest point (ICP) based optimization, yet they rely on an accurate pose prior to avoid local minima. Yu et al.~\cite{depth-shift-2} studied the Sim(3) registration within the depth priors-based two-view geometry problem, which can be extended to correspondence pruning. In summary, despite the progress achieved by existing depth prior-based methods \cite{graph-i2p-cvpr, mast3r-slam, depth-shift-2}, they fail to adequately handle the depth noise during Sim(3) registration. Therefore, it is crucial to develop the novel pruning mechanism that are robust against noisy depth priors.


\section{Pruning as an Extended Sim(3) Registration Problem}
\label{sec.task}

\begin{table}[t]
\setlength{\tabcolsep}{2pt}
\centering
\caption{
Comparison of SE(3), Sim(3), and extended Sim(3) registration formulations. In extended Sim(3), each correspondence is associated with an individual scale factor $s_i$, increasing the number of unknowns from N+7 to 2N+6, which renders the problem underdetermined without additional structural constraints.
}
\resizebox{1.0\linewidth}{!}{
\begin{tabular}{l|c|c|c}
\toprule
Registration problem & Inlier constraint & Objectives& Num. of variables  \\
\midrule
SE(3)   & $\mathbf{p}_i=\mathbf{R}\mathbf{q}_i+\mathbf{t}$ & Estimate inliers $\{w_i\in\{0,1\}\}_{i=1}^\mathrm{N}$, $\mathbf{R}$, $\mathbf{t}$ & $\mathrm{N}+6$ \\
Regular Sim(3) & $\,\,\,\mathbf{p}_i=s\mathbf{R}\mathbf{q}_i+\mathbf{t}$ & Estimate inliers $\{w_i\in\{0,1\}\}_{i=1}^\mathrm{N}$, $\mathbf{R}$, $\mathbf{t}$, $s$ & $\mathrm{N}+7$ \\
\rowcolor{gray!20} Extended Sim(3) & $\,\,\mathbf{p}_i=s_i\mathbf{R}\mathbf{q}_i+\mathbf{t}$ & Estimate inliers $\{w_i\in\{0,1\}\}_{i=1}^\mathrm{N}$, $\mathbf{R}$, $\mathbf{t}$, $\{s_i\}_{i=1}^\mathrm{N}$ & $2\mathrm{N}+6$ \\
\bottomrule
\end{tabular}}
\label{table:problem_setting}
\vspace{-12pt}
\end{table}

We motivate the necessity of framing pruning as an extended Sim(3) registration.

\subsection{Problem Formulation}
\label{sec.task.A}

Given 2D-3D correspondences $\{\mathbf{c}_i\}_{i=1}^{\mathrm{N}}$ predicted from a learning-based I2P registration model~\cite{2d3d-matr}, correspondence pruning seeks to filter out outlier matches from $\{\mathbf{c}_i\}_{i=1}^{\mathrm{N}}$. 
Each correspondence $\mathbf{c}_i\triangleq(\mathbf{l}_i, \mathbf{p}_i)$ consists of a 2D pixel coordinate  $\mathbf{l}_i=({u}_i,{v}_i,1)^T\in\mathbb{R}^3$ and a 3D point coordinate $\mathbf{p}_i=({x}_i, {y}_i, {z}_i)^T\in\mathbb{R}^3$. A correspondence $\mathbf{c}_i$ is an outlier if it does not satisfy the following constraint: 
\begin{equation}
\label{eq_1}
\Vert \mathbf{l}_i - \pi(\mathbf{p}_i| \mathbf{K},\mathbf{R}, \mathbf{t}) \Vert_2 \leq \delta_{\mathrm{2d}}
\end{equation}
\noindent where $\pi(\cdot)$ is the standard camera projection function~\cite{camera_model}. $\mathbf{K}\in\mathbb{R}^{3\times 3}$ denotes the camera intrinsic matrix. $\mathbf{R}\in \mathrm{SO}(3)$ and $\mathbf{t}\in\mathbb{R}^{3\times 1}$ represent the camera pose. $\delta_{\mathrm{2d}}$ is a reprojection error threshold. In essence, 2D-3D correspondence pruning outputs inliers and camera pose. 

\subsection{Pruning as a Regular Sim(3) Registration}
\label{sec.task.B}

Current methods formulate the pruning as a regular Sim(3) registration problem \cite{graph-i2p-cvpr, depth-shift-2, mast3r-slam}. Using the pre-trained visual foundation models~\cite{depthanything, dep_anything_v2}, we estimate a metric depth from a monocular RGB image. 3D-3D correspondence $\mathbf{d}_i\triangleq(\mathbf{q}_i, \mathbf{p}_i)$ is then obtained by lifting $\mathbf{c}_i$ with depth priors. $\mathbf{q}_i = d_i \cdot \pi^{-1}(\mathbf{l}_i|\mathbf{K})$
where $\mathbf{l}_i$ and $d_i$ denote the pixel coordinate and predicted depth of the $i$-th pixel. $\pi^{-1}(\cdot)$ is the camera back-projection function~\cite{camera_model}. Then, pruning is formulated as a regular Sim(3) registration problem (summarized in Table \ref{table:problem_setting}):
\begin{equation}
\label{eq_reg_sim_3}
\min_{\{w_i\}_{i=1}^\mathrm{N},s,\mathbf{R}, \mathbf{t}} \sum_{i=1}^\mathrm{N} w_i\Vert \mathbf{p}_i - (s\cdot \mathbf{R} \mathbf{q}_i + \mathbf{t}) \Vert_2^2
\end{equation}
where $s\in\mathbb{R}^+$ is the scale of depth priors. $w_i\in\{0,1\}$ denotes whether the $i$-th correspondence is inlier or outlier. In general, Sim(3) registration can be solved within two steps: (i) scale estimation and (ii) pose estimation~\cite{TEASER}. The first step targets estimating the scale factor $s$. By employing the constraint in Eq.~\ref{eq_4}, the scale 
$s$ is estimated via the adaptive voting algorithm~\cite{TEASER}.
\begin{equation}
\label{eq_4}
s = \Vert \mathbf{p}_i - \mathbf{p}_j \Vert_2 /\Vert \mathbf{q}_i - \mathbf{q}_j \Vert_2 , \,\, \mathrm{if\,\, \mathbf{c}_i, \mathbf{c}_j \,\,are\,\, inliers}.
\end{equation}
After scale estimation, Sim(3) registration is reduced to an SE(3) registration. The second step involves estimating $\{w_i\}_{i=1}^\mathrm{N}$, $\mathbf{R}$ and $\mathbf{t}$, which can be effectively solved via  compatibility graph based SE(3) registration methods~\cite{mac, sc2-pcr, fastmac, turboreg}. Overall, the regular Sim(3) registration formulated in Eq.~\ref{eq_reg_sim_3} can be fully addressed using existing techniques.

\subsection{From Regular Sim(3) to Extended Sim(3)}
\label{sec.task.C}

Actually, Eq.~\ref{eq_reg_sim_3} exhibits a \textbf{critical limitation} in that it does not fully account for the influence of depth noise. Due to scale ambiguity in monocular images, depth priors obtained even from well-trained visual foundation models still suffer from significant errors~\cite{depth-shift-1, depth-shift-2, depth-shift-3}. In this paper, we model the depth noise by extending the original Sim(3) registration problem. Instead of using affine correction models \cite{depth-shift-1, depth-shift-3}, we consider the multiplicative depth noise term $\delta_i$ on $d_i$\footnote{multiplicative noise can be equivalently represented as additive noise.} and derive 
$\delta_i\mathbf{q}_i = \delta_i d_i \cdot \pi^{-1}(\mathbf{l}_i|\mathbf{K})$. Then, Eq. \ref{eq_reg_sim_3} is extended as a more general registration problem:
\begin{equation}
\label{eq_reg_sim_3_ex}
\min_{\{w_i\}_{i=1}^\mathrm{N},\{s_i\}_{i=1}^\mathrm{N},\mathbf{R}, \mathbf{t}} \sum_{i=1}^\mathrm{N} w_i\Vert \mathbf{p}_i - (s_i\cdot \mathbf{R} \mathbf{q}_i + \mathbf{t}) \Vert_2^2
\end{equation}
where $s_i=s \cdot \delta_i$. We call Eq. \ref{eq_reg_sim_3_ex} \textbf{extended Sim(3) registration} because every inlier follows a slightly different similarity transformation parametrized by $s_i$, $\mathbf{R}$, and $\mathbf{t}$. Compared to regular Sim(3) registration, extended Sim(3) registration explicitly optimizes for depth noise, leading to more stable I2P registration. 

\section{Pruning with Extended Sim(3) Registration}
\label{sec.method}

To solve the extended Sim(3) registration, we design a heuristic yet efficient pruning algorithm called \textbf{Ex-Sim(3)-Reg} with an in-depth theoretical analysis.  

\subsection{Extended Compatibility Graph}
\label{sec.method.P}

Compatibility graph is a fundamental structure in SE(3) registration \cite{sc2-pcr, mac, fastmac}. Inliers can be identified by searching the maximum clique in this graph \cite{mac}. We extend this concept to solve Eq. \ref{eq_reg_sim_3_ex}. Given the lifted 3D-3D correspondences $\{\mathbf{d}_i\}_{i=1}^{\mathrm{N}}$, we define the first-order extended compatibility graph as $\mathcal{G}_{\textrm{FOG}}(\zeta)$. It consists of $\mathrm{N}$ vertices and each vertex denotes a correspondence $\mathbf{d}_i$. $\zeta$ is a scale parameter. The adjacency matrix $\mathbf{W}_{\textrm{FOG}}(\zeta)\in \mathbb{R}^{\mathrm{N}\times \mathrm{N}}$ describes the edge connections of $\mathcal{G}_{\textrm{FOG}}(\zeta)$~\cite{mac}:
\begin{equation}
\label{eq_fog_ex_1}
\mathbf{W}_{\textrm{FOG}}(\zeta)_{ij} = 
\begin{cases}
1, \,\, d_{ij}(\zeta) \leq d_{thr} \\
0, \,\, d_{ij}(\zeta) > d_{thr}
\end{cases}, d_{ij}(\zeta) = \vert \Vert \mathbf{p}_i-\mathbf{p}_j \Vert_2 - \Vert \zeta\cdot(\mathbf{q}_i - \mathbf{q}_j) \Vert_2 \vert
\end{equation}
where $d_{thr}\geq0$ is a distance threshold. For a matrix $\mathbf{W}$, $\mathbf{W}_{ij}$ is the element in the $i$-th row and $j$-th column. We further construct the second-ordered extended compatibility graph as $\mathcal{G}_{\textrm{SOG}}(\zeta)$, which shares the same vertex set as $\mathcal{G}_{\textrm{FOG}}(\zeta)$ but with a different adjacency matrix $\mathbf{W}_{\textrm{SOG}}(\zeta)$ \cite{sc2-pcr}:
\begin{equation}
\label{eq_sog_ex_1}
\mathbf{W}_{\textrm{SOG}}(\zeta) = 
 \mathbf{W}_{\textrm{FOG}}(\zeta) \circ
 (\mathbf{W}_{\textrm{FOG}}(\zeta) \cdot\mathbf{W}_{\textrm{FOG}}(\zeta))
\end{equation}
\noindent where $\circ$ denotes element-wise multiplication. We name $\mathcal{G}_{\textrm{FOG}}(\zeta)$ and $\mathcal{G}_{\textrm{SOG}}(\zeta)$ as the extended compatibility graphs, since their edge relations are parameterized by a scale variable $\zeta$. As second-order graphs have been empirically validated for the 3D-3D registration~\cite{sc2-pcr}, we adopt $\mathcal{G}_{\textrm{SOG}}(\zeta)$ in the following pruning method. 

\subsection{Pruning based on Extended Compatibility Graphs}
\label{sec.method.A}


\begin{figure}[t]
	\centering
		\includegraphics[width=1.0\linewidth]{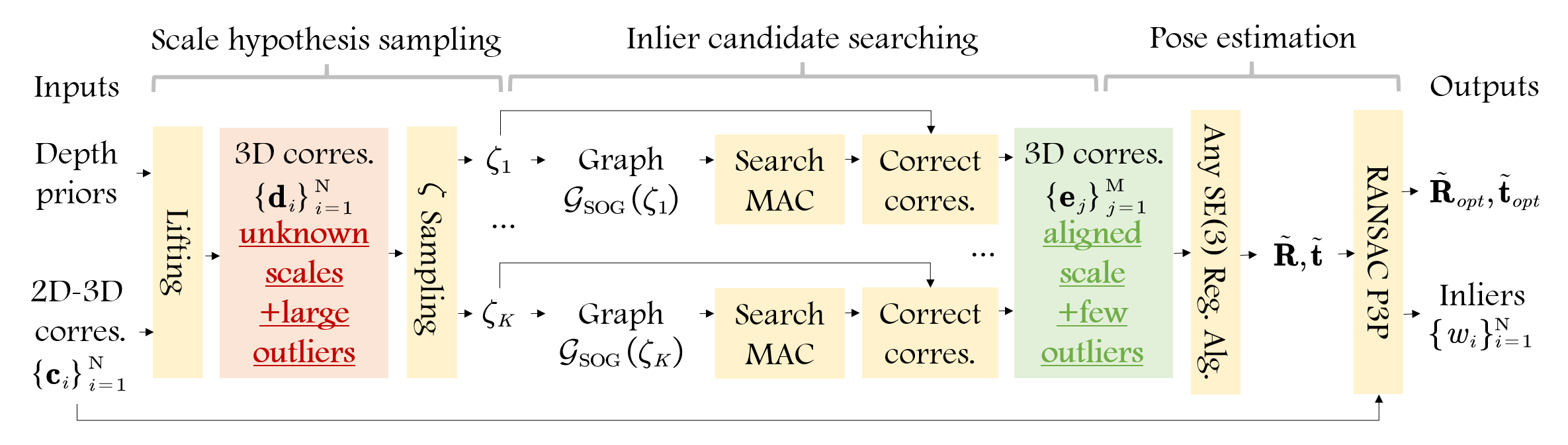}	
		\caption{Pipeline of Ex-Sim(3)-Reg. Through scale hypothesis sampling and inlier candidate search, 3D correspondences $\{\mathbf{d}_i\}_{i=1}^\mathrm{N}$ are transformed to $\{\mathbf{e}_j\}_{j=1}^\mathrm{M}$. This reduces the extended Sim(3) registration into an SE(3) registration problem, enabling robust inlier estimation using off-the-shelf SE (3) solvers.}
	\label{fig:pipeline}
    \vspace{-12pt}
\end{figure}

\vspace{+1mm}
\noindent\textbf{Pruning prototype}. Extended Sim(3) registration is an underdetermined problem and cannot be solved directly. To tackle this challenge, our pruning prototype aims to \textbf{approximate} the \textbf{extended Sim(3)} registration as a \textbf{standard SE(3)} registration by scale-correcting correspondences using a family of graphs $\{\mathcal{G}_{\textrm{SOG}}(\zeta)\}_\zeta$, as illustrated in Fig. \ref{fig:pipeline}. 

Specifically, we first analyze the maximum clique (MAC) of the extended compatibility graph, denoted as $\mathrm{Mac}(\mathcal{G}_{\textrm{SOG}}(\zeta))$. Under the inlier set maximization assumption \cite{bnb-blind-pnp-1,mac}, $\mathrm{Mac}(\mathcal{G}_{\textrm{SOG}}(\zeta))$ contains potential inliers. From the definition of extended compatibility graph in Eq. \ref{eq_fog_ex_1}, if $\mathbf{d}_i=(\mathbf{q}_i, \mathbf{p}_i) \in \mathrm{Mac}(\mathcal{G}_{\textrm{SOG}}(\zeta))$, $\mathbf{e}_i\triangleq(\zeta\cdot\mathbf{q}_i, \mathbf{p}_i)$ is an inlier candidate obeying the SE(3) transformation ($\mathbf{p}_i=\mathbf{R}(\zeta\mathbf{q}_i)+\mathbf{t}$). We write $\mathbf{e}_i=\varphi(\mathbf{d}_i,\zeta)$ and $\mathbf{e}_i \in \varphi(\mathrm{Mac}(\mathcal{G}_{\textrm{SOG}}(\zeta)), \zeta)$ where $\varphi(\cdot)$ is called the scale-correction function. All such $\mathbf{e}_i$ yield a new correspondence set:
\begin{equation}
\label{eq_set_all}
\{\mathbf{e}_j\}_{j=1}^\mathrm{M} = \cup_{\zeta\in(0,+\infty)} \varphi(\mathrm{Mac}(\mathcal{G}_{\textrm{SOG}}(\zeta)), \zeta)
\end{equation}

Using function $\varphi(\cdot)$, the extended Sim(3) is converted as an SE(3) registration problem over $\{\mathbf{e}_j\}_{j=1}^\mathrm{M}$. The set $\{\mathbf{e}_j\}_{j=1}^\mathrm{M}$ has two advantages: (i) through maximum clique searching, {inlier ratio} of $\{\mathbf{e}_j\}_{j=1}^\mathrm{M}$ is \textbf{higher than} $\{\mathbf{d}_i\}_{i=1}^\mathrm{N}$; (ii) the inlier constrain in $\{\mathbf{e}_j\}_{j=1}^\mathrm{M}$ is simplified as an SE(3) transformation, which is \textbf{easier} to solve than the extended Sim(3) transformation. 

Once $\{\mathbf{e}_j\}_{j=1}^\mathrm{M}$ are obtained, camera pose and inliers are robustly estimated:
\begin{equation}
\label{eq_proto_prune}
\mathbf{\tilde{R}}, \mathbf{\tilde{t}} = \mathrm{SE(3)\_Reg} (\{\mathbf{e}_j\}_{j=1}^\mathrm{M})
\end{equation}
where $\mathbf{\tilde{R}}, \mathbf{\tilde{t}}$ are estimated camera pose. $\mathrm{SE(3)\_Reg}(\cdot)$ stands for any off-the-shelf SE(3) registration algorithm. With $\mathbf{\tilde{R}}$ and $\mathbf{\tilde{t}}$, inlier labels $\{w_i\}_{i=1}^\mathrm{N}$ are identified by Eq. \ref{eq_1}. Thus, the proposed pruning prototype (Eqs. \ref{eq_set_all} and \ref{eq_proto_prune}) provides a feasible solution to Eq. \ref{eq_reg_sim_3_ex}. 

\vspace{+1mm}
\noindent\textbf{Pruning algorithm}. Built upon the above pruning prototype, we propose an efficient algorithm {Ex-Sim(3)-Reg} with three steps: (i) scale hypothesis sampling, (ii) inlier candidate searching, and (iii) pose estimation (Fig. \ref{fig:pipeline}). The \textbf{first step} is to sample $\zeta$, 
since exhaustive search over $\zeta\in(0,+\infty)$ is infeasible. From Eq.~\ref{eq_4}, a coarse $\tilde{s}$ is estimated by the adaptive voting algorithm~\cite{TEASER}. We heuristically restrict the search range to $\zeta \in (\tilde{s}-\Delta s, \tilde{s}+\Delta s)$ where $\Delta s$ is a hyperparameter. We then uniformly sample $K$ scales $\{\zeta_k\}_{k=1}^K$ from this interval and build the corresponding graphs $\{\mathcal{G}_{\textrm{SOG}}(\zeta_k))\}_{k=1}^K$ via Eqs. \ref{eq_fog_ex_1} and \ref{eq_sog_ex_1}. The \textbf{second step} is to construct $\{\mathbf{e}_j\}_{j=1}^\mathrm{M}$ from Eq. \ref{eq_set_all} where $\zeta \in (0,+\infty)$ is replaced by $\{\zeta_k\}_{k=1}^K$. The \textbf{third step} is to estimate pose and inliers. Camera pose is first estimated from Eq. \ref{eq_proto_prune} and then refined with RANSAC based P3P algorithm \cite{ransac-p3p}:
\begin{equation}
\label{eq_post_prune}
\mathbf{\tilde{R}}_{opt}, \mathbf{\tilde{t}}_{opt}, \{w_i\}_{i=1}^\mathrm{N} = \mathrm{RANSAC\_P3P}(\{\mathbf{c}_i\}_{i=1}^{\mathrm{N}}, \mathbf{\tilde{R}}, \mathbf{\tilde{t}})
\end{equation}

\noindent In Eq. \ref{eq_post_prune}, given the reliable pose $\mathbf{\tilde{R}}$ and $\mathbf{\tilde{t}}$ as priors, RANSAC based P3P \cite{ransac-p3p} first pre-filters the outliers with the reprojection error larger than $\delta_{thr}$ and then estimate the more accurate pose $\mathbf{\tilde{R}}_{opt}, \mathbf{\tilde{t}}_{opt}$ using the remained inliers. The ablation study of $\delta_{thr}$ is provided in Sec. \ref{sec.exp.C}. More details of the algorithm are presented in the supplemental materials.  

\subsection{Theoretical Analysis of Pruning Prototype}
\label{sec.method.B}


To further justify the effectiveness of our method, we introduce the following key concept:


\vspace{+1mm}
\noindent\textbf{Ideal extended compatibility graph}. If $\{s_i\}_{i=1}^{\mathrm{N}}$ in Eq. \ref{eq_reg_sim_3_ex} are known, we could construct the ideal extended compatibility graphs $\mathcal{G}_{\textrm{FOG}}(\{s_i\}_{i=1}^{\mathrm{N}})$, $\mathcal{G}_{\textrm{SOG}}(\{s_i\}_{i=1}^{\mathrm{N}})$. Edge matrix $\mathbf{W}_{\textrm{FOG}}(\{s_i\}_{i=1}^{\mathrm{N}})$ is defined analogously to Eq. \ref{eq_fog_ex_1} where $d_{ij}(\zeta)$ is replaced by $\rho_{ij}(s_i,s_j) = \vert \Vert \mathbf{p}_i-\mathbf{p}_j \Vert_2 - \Vert s_i\cdot\mathbf{q}_i - s_j\cdot\mathbf{q}_j \Vert_2 \vert$. Similar to Eq. \ref{eq_sog_ex_1}, $\mathbf{W}_{\textrm{SOG}}(\{s_i\}_{i=1}^{\mathrm{N}})$ is then computed as \cite{sc2-pcr}:
\begin{equation}
\label{eq_sog_ex_1_ideal}
\mathbf{W}_{\textrm{SOG}}(\{s_i\}_{i=1}^{\mathrm{N}}) = 
\mathbf{W}_{\textrm{FOG}}(\{s_i\}_{i=1}^{\mathrm{N}}) \circ (\mathbf{W}_{\textrm{FOG}}(\{s_i\}_{i=1}^{\mathrm{N}})\mathbf{W}_{\textrm{FOG}}(\{s_i\}_{i=1}^{\mathrm{N}}))
\end{equation}
Under the inlier set maximization assumption \cite{bnb-blind-pnp-1,mac}, solving the extended Sim(3) registration in Eq. \ref{eq_reg_sim_3_ex} can be interpreted as searching for the maximum clique in the ideal extended compatibility graph $\mathcal{G}_{\textrm{SOG}}(\{s_i\}_{i=1}^{\mathrm{N}})$.




\vspace{+1mm}
\noindent\textbf{Relation between ideal and practical graphs}. Based on the above, we investigate the relationship between the ideal graph $\mathcal{G}_{\textrm{SOG}}(\{s_i\}_{i=1}^{\mathrm{N}})$ and the practical graphs $\{\mathcal{G}_{\textrm{SOG}}(\zeta_k)\}_{k=1}^K$. Let $\mathscr{F}$ denote the ground truth inlier set. We define a subset $\mathscr{F}_k \subseteq \mathscr{F}$ that if $\mathbf{d}_i$ and $\mathbf{d}_j \in \mathscr{F}_k$, there exists a unique $\zeta_k$ satisfying $\rho_{ij}(\zeta_k) \leq d_{thr} < \rho_{ij}(\zeta_l)$ if $l\neq k$. Under the inlier set maximization assumption \cite{bnb-blind-pnp-1,mac}, we have $\mathscr{F}_k  \subseteq \mathrm{Mac}(\mathcal{G}_{\textrm{SOG}}(\zeta_k))$ and $\mathscr{F} \subseteq\mathrm{Mac}(\mathcal{G}_{\textrm{SOG}}(\{s_i\}_{i=1}^{\mathrm{N}}))$. This leads to the following relation: 
\begin{equation}
\label{eq_relation_a}
\cup_{k=1}^{K} \mathscr{F}_k \subseteq \mathrm{Mac}(\mathcal{G}_{\textrm{SOG}}(\{s_i\}_{i=1}^{\mathrm{N}})) \cap \left(\cup_{k=1}^{K}\mathrm{Mac}(\mathcal{G}_{\textrm{SOG}}(\zeta_k))\right)
\end{equation} 

This shows that the maximum cliques of ideal and practical graphs overlap on $\cup_{k=1}^K \mathscr{F}_k$. In short, the union of practical graphs $\{\mathcal{G}_{\textrm{SOG}}(\zeta_k)\}_{k=1}^K$ serves as an approximation of the ideal graph $\mathcal{G}_{\textrm{SOG}}(\{s_i\}_{i=1}^{\mathrm{N}})$, which theoretically validates Ex-Sim(3)-Reg. In addition, Eq. \ref{eq_relation_a} reveals that our solution only searches for inliers within $\cup_{k=1}^K \mathscr{F}_k$, while the various experiments in Sec. \ref{sec.exp.B} verify that this subset is sufficient to achieve robust pruning in most real-world scenarios. 


\begin{figure}[t]
	\centering
		\includegraphics[width=1.0\linewidth]{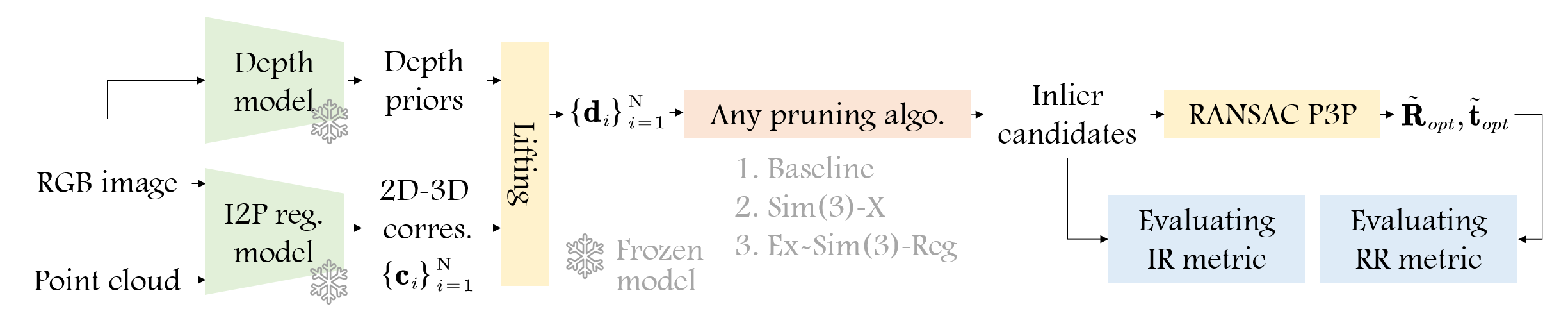}	
		\caption{Evaluation protocol. Inlier Ratio (IR) measures the quality of inlier candidates, while Registration Recall (RR) evaluates the accuracy of the estimated rotation $\mathbf{\tilde{R}}_{opt}$ and translation $\mathbf{\tilde{t}}_{opt}$. \textbf{Baseline} denotes {no pruning}. \textbf{Sim(3)-X} estimates a global scale $\tilde{s}$~\cite{TEASER}, transforms correspondences from $\{\mathbf{d}_i\}_{i=1}^\mathrm{N}$ to $\{(\tilde{s}\mathbf{q}_i, \mathbf{p}_i)\}_{i=1}^\mathrm{N}$, and then applies an SE(3) solver \textbf{X} to identify inliers. A complete protocol is in the supplemental materials.}
	\label{fig:eval}
    \vspace{-6pt}
\end{figure}

\begin{table}[t]
\setlength{\tabcolsep}{2pt}
\centering
\caption{
Correspondence pruning performance on the 7-Scenes dataset~\cite{scene-7-dataset} under the disturbed condition. 
\textbf{Bold} indicates the best mean performance across scenes.
}
\begin{tabular}{l|ccccccc}
\toprule
Method & SC-1 & SC-2 & SC-3 & SC-4 & SC-5 &  SC-6 & Mean \\
\midrule
\multicolumn{8}{c}{\emph{Inlier Ratio} $\uparrow$} \\
\midrule
Baseline & 0.295 & 0.257 & 0.325 & 0.265 & 0.227 & 0.228 & 0.266 \\
Sim(3)-FastMAC  & 0.374 & 0.316 & 0.379 & 0.328 & 0.296 & 0.306 & 0.333 \\
Sim(3)-SC2-PCR  & 0.481 & 0.435 & 0.431 & 0.446 & 0.436 & 0.475 & 0.451 \\
Sim(3)-SC2-PCR++  & 0.295 & 0.254 & 0.325 & 0.259 & 0.229 & 0.251 & 0.268 \\
Sim(3)-MAC      & 0.372 & 0.313 & 0.365 & 0.320 & 0.307 & 0.322 & 0.333 \\
Sim(3)-TurboReg & 0.369 & 0.311 & 0.373 & 0.316 & 0.306 & 0.323 & 0.333 \\
\rowcolor{gray!20} Ex-Sim(3)-Reg     & \textbf{0.576} & \textbf{0.553} & \textbf{0.601} & \textbf{0.563} & \textbf{0.487} & \textbf{0.510} & \textbf{0.548} (\textbf{+28.2\%})\\
\midrule
\multicolumn{8}{c}{\emph{Registration Recall} $\uparrow$} \\
\midrule
Baseline & 0.385 & 0.247 & 0.083 & 0.255 & 0.429 & 0.357 & 0.292 \\
Sim(3)-FastMAC  & 0.246 & 0.192 & 0.251 & 0.170 & 0.411 & 0.429 & 0.283 \\
Sim(3)-SC2-PCR  & 0.323 & 0.247 & 0.250 & 0.319 & 0.369 & 0.143 & 0.275 \\
Sim(3)-SC2-PCR++  & 0.046 & 0.137 & 0.083 & 0.043 & 0.060 & 0.071 & 0.073 \\
Sim(3)-MAC      & 0.338 & 0.260 & 0.167 & 0.234 & 0.440 & 0.214 & 0.275 \\
Sim(3)-TurboReg & 0.400 & 0.327 & 0.170 & 0.298 & 0.435 & 0.429 & 0.343 \\
\rowcolor{gray!20} Ex-Sim(3)-Reg     & \textbf{0.415} & \textbf{0.493} & \textbf{0.500} & \textbf{0.383} & \textbf{0.542} & \textbf{0.500} & \textbf{0.472} (\textbf{+18.0\%}) \\
\bottomrule
\end{tabular}
\vspace{-5pt}
\label{table:att_res}
\end{table}

\section{Experiments and Discussions}
\label{sec.exp}

\subsection{Experimental Setup} 
\label{sec.exp.A}

To evaluate the accuracy of 2D-3D correspondence pruning, we conduct a series of experiments on the public indoor datasets, including 7-Scenes~\cite{scene-7-dataset}, RGBD-v2~\cite{RGBD-dataset}, ScanNet~\cite{scan-net}, and TUM~\cite{tum}. To prepare the pruning dataset, we adopt a pre-trained I2P registration model MATR~\cite{2d3d-matr} to generate 2D-3D correspondences under three challenging settings (disturbed, unseen, and low-inlier conditions). More details of these settings are provided in the supplemental materials. 

For fair comparison, our method is compared with the state-of-the-art non-learning correspondence pruning methods across four datasets, where the evaluation procedure is provided in Fig. \ref{fig:eval}. \textbf{Baseline} is the default pruning method. Pruning method with regular Sim(3) registration is marked as \textbf{Sim(3)-X} where \textbf{X} denotes the arbitrary SE(3) registration algorithm, like SC2-PCR~\cite{sc2-pcr}, SC2-PCR++~\cite{sc2-pcr++}, MAC~\cite{mac}, FastMAC~\cite{fastmac}, and TurboReg~\cite{turboreg}. In the depth priors based methods, depths are predicted from RGB images using Depth Anything v2 (ViT-Large)~\cite{dep_anything_v2}. Implementation details and hyperparameter configurations of the proposed method are provided in the supplemental materials. $\mathbf{{R}}_{gt}$ and $\mathbf{{t}}_{gt}$ are the ground truth camera pose. Two metrics are used for evaluation (computation details refer to the appendix in \cite{2d3d-matr}):
\begin{itemize}
    \item Inlier Ratio (IR) evaluates the ratio of inliers from inlier candidates as shown in Fig. \ref{fig:eval}. A 2D-3D correspondence $(\mathbf{l}_i,\mathbf{p}_i)$ can be regarded as inlier if the 3D distance between points $\mathbf{{R}}_{gt}\mathbf{p}_i+\mathbf{{t}}_{gt}$ and $d_{i,gt}\cdot\pi^{-1}(\mathbf{l}_i|\mathbf{K})$ ($d_{i,gt}$ is the ground truth depth of the pixel $\mathbf{l}_i$) is below a default threshold of 10cm. 
    \item Registration Recall (RR) evaluates the accuracy of $\mathbf{\tilde{R}}_{opt}, \mathbf{\tilde{t}}_{opt}$. A successful registration is identified if the average 3D distance between 3D point clouds transformed by $(\mathbf{\tilde{R}}_{opt}, \mathbf{\tilde{t}}_{opt})$ and $(\mathbf{{R}}_{gt}, \mathbf{{t}}_{gt})$ is less than a default threshold of 5cm.
\end{itemize}


\begin{table}[t]
\setlength{\tabcolsep}{2pt}
\centering
\caption{
Correspondence pruning performance on the TUM dataset~\cite{tum} under the disturbed condition.
 \textbf{Bold} indicates the best mean performance.
}
\begin{tabular}{l|cccc}
\toprule
Method & Freiburg\_01 & Freiburg\_02 & Freiburg\_03 & Mean \\
\midrule
\multicolumn{5}{c}{\emph{Inlier Ratio} $\uparrow$} \\
\midrule
Baseline & 0.242 & 0.265 & 0.183 & 0.230 \\
Sim(3)-FastMAC  & 0.317 & 0.363 & 0.241 & 0.307 \\
Sim(3)-SC2-PCR  & 0.387 & 0.415 & 0.331 & 0.378 \\
Sim(3)-SC2-PCR++  & 0.267 & 0.339 & 0.235 & 0.280  \\
Sim(3)-MAC      & 0.323 & 0.355 & 0.231 & 0.303  \\
Sim(3)-TurboReg & 0.316 & 0.347 & 0.250 & 0.304  \\
\rowcolor{gray!20} Ex-Sim(3)-Reg     & \textbf{0.428} & \textbf{0.548} & \textbf{0.480} & \textbf{0.485} (\textbf{+25.5\%})  \\
\midrule
\multicolumn{5}{c}{\emph{Registration Recall} $\uparrow$} \\
\midrule
Baseline & 0.160 & 0.040 & 0.152 & 0.117 \\
Sim(3)-FastMAC  & 0.092 & 0.037 & 0.121 & 0.083 \\
Sim(3)-SC2-PCR  & 0.067 & 0.037 & 0.128 & 0.077 \\
Sim(3)-SC2-PCR++  & 0.050 & 0.029 & 0.058 & 0.046  \\
Sim(3)-MAC      & 0.109 & 0.040 & 0.112 & 0.087  \\
Sim(3)-TurboReg & 0.118 & 0.049 & 0.125 & 0.097  \\
\rowcolor{gray!20} Ex-Sim(3)-Reg     & \textbf{0.277} & \textbf{0.198} & \textbf{0.432} & \textbf{0.302} (\textbf{+18.5\%})  \\
\bottomrule
\end{tabular}
\vspace{-5pt}
\label{table:att_res_ex}
\end{table}

\subsection{Main Comparisons} 
\label{sec.exp.B}

\vspace{+1mm}
\noindent\textbf{Performances under disturbed condition}. We evaluate the compared methods in the stress-test scenarios. We simulate the disturbed condition and collect the dataset in two steps: (1) distorting color features of point cloud by the common adversarial attack algorithm \cite{attack_point_cloud}; (2) generating correspondences for I2P pruning by inferring MATR \cite{2d3d-matr} with RGB images and distorted point clouds. Results on the 7-Scenes dataset \cite{scene-7-dataset} are provided in Table \ref{table:att_res}, where MATR \cite{2d3d-matr} was pre-trained on the Chess scene and generated the pruning dataset on the other six scenes. Our method improves IR by \textbf{28.2\%} and RR by \textbf{18.0\%} compared to the baseline. Consistent improvements are observed on the TUM dataset~\cite{tum} (Table~\ref{table:att_res_ex}), where our approach achieves substantial RR improvements ($\textbf{20\%}$) on Freiburg\_02, while all baselines remain below $5\%$. This demonstrates strong robustness against feature corruption. Results with the different depth prediction methods are provided in the supplemental materials.

\begin{table}[t]
\setlength{\tabcolsep}{2pt}
\centering
\caption{
Cross-dataset generalization from 7-Scenes~\cite{scene-7-dataset} to RGBD-v2~\cite{RGBD-dataset}. \textbf{Bold} indicates the best mean performance. Results demonstrate robustness to domain shift in depth priors and correspondence distributions.
}
\resizebox{1\linewidth}{!}{
\begin{tabular}{l|cccccccc}
\toprule
Method & RG-1 & RG-2 & RG-3 & RG-4 & RG-5 & RG-6 & RG-7 & Mean \\
\midrule
\multicolumn{9}{c}{\emph{Inlier Ratio} $\uparrow$} \\
\midrule
Baseline & 0.242 & 0.248 & 0.223 & 0.224 & 0.273 & 0.282 & 0.259 & 0.250 \\
Sim(3)-FastMAC  & 0.309 & 0.326 & 0.275 & 0.283 & 0.354 & 0.341 & 0.350 & 0.320 \\
Sim(3)-SC2-PCR  & 0.413 & 0.415 & 0.352 & 0.365 & 0.367 & 0.369 & 0.373 & 0.379 \\
Sim(3)-SC2-PCR++  & 0.205 & 0.252 & 0.239 & 0.231 & 0.309 & 0.322 & 0.248 & 0.258 \\
Sim(3)-MAC      & 0.322 & 0.317 & 0.281 & 0.281 & 0.307 & 0.319 & 0.308 & 0.305 \\
Sim(3)-TurboReg & 0.391 & 0.315 & 0.280 & 0.279 & 0.307 & 0.321 & 0.306 & 0.314 \\
\rowcolor{gray!20} Ex-Sim(3)-Reg     & \textbf{0.466} & \textbf{0.455} & \textbf{0.452} & \textbf{0.428} & \textbf{0.471} & \textbf{0.472} & \textbf{0.442} & \textbf{0.455} (\textbf{+20.5\%}) \\
\midrule
\multicolumn{9}{c}{\emph{Registration Recall} $\uparrow$} \\
\midrule
Baseline & 0.237 & 0.240 & 0.226 & 0.444 & 0.267 & 0.159 & 0.188 & 0.252 \\
Sim(3)-FastMAC  & 0.263 & 0.280 & 0.226 & 0.407 & 0.173 & 0.159 & 0.167 & 0.239 \\
Sim(3)-SC2-PCR  & 0.263 & 0.320 & 0.355 & 0.333 & \textbf{0.280} & 0.058 & 0.188 & 0.257 \\
Sim(3)-SC2-PCR++  & 0.026 & 0.031 & 0.055 & 0.061 & 0.253 & 0.101 & \textbf{0.229} & 0.108 \\
Sim(3)-MAC      & 0.342 & 0.201 & 0.323 & 0.370 & 0.200 & 0.043 & 0.104 & 0.226 \\
Sim(3)-TurboReg & 0.289 & 0.240 & 0.324 & 0.369 & 0.160 & 0.232 & 0.188 & 0.257 \\
\rowcolor{gray!20} Ex-Sim(3)-Reg     & \textbf{0.500} & \textbf{0.320} & \textbf{0.484} & \textbf{0.556} & \textbf{0.280} & \textbf{0.333} & {0.208} & \textbf{0.383} (\textbf{+13.1\%}) \\
\bottomrule
\end{tabular}}
\vspace{-5pt}
\label{table:cross_res_1}
\end{table}

\begin{table}[!ht]
\setlength{\tabcolsep}{2pt}
\centering
\caption{
Cross-dataset evaluation from 7-Scenes~\cite{scene-7-dataset} to ScanNet~\cite{scan-net}. `N/A' indicates failure to return a result within 1 second, reflecting computational instability under severe outlier contamination.
}
\resizebox{1\linewidth}{!}{
\begin{tabular}{l|cccccccc}
\toprule
Method & SN-1 & SN-2 & SN-3 & SN-4 & SN-5 & SN-6 & SN-7 & Mean \\
\midrule
\multicolumn{9}{c}{\emph{Inlier Ratio} $\uparrow$} \\
\midrule
Baseline & 0.463 & 0.469 & 0.388 & 0.269 & 0.474 & 0.408 & 0.390 & 0.409 \\
Sim(3)-FastMAC  & 0.553 & 0.570 & 0.478 & 0.344 & 0.566 & 0.501 & 0.566 & 0.511 \\
Sim(3)-SC2-PCR  & 0.625 & 0.616 & 0.552 & 0.398 & 0.629 & 0.594 & 0.644 & 0.579 \\
Sim(3)-SC2-PCR++  & 0.473 & 0.487 & 0.377 & 0.297 & 0.463 & 0.442 & 0.472 & 0.430 \\
Sim(3)-MAC      & N/A & N/A & N/A & N/A & N/A & N/A & N/A & N/A \\
Sim(3)-TurboReg & 0.517 & 0.506 & 0.434 & 0.338 & 0.521 & 0.482 & 0.520 & 0.474 \\
\rowcolor{gray!20} Ex-Sim(3)-Reg     & \textbf{0.694} & \textbf{0.747} & \textbf{0.674} & \textbf{0.423} & \textbf{0.743} & \textbf{0.639} & \textbf{0.715} & \textbf{0.662} (\textbf{+25.3\%}) \\
\midrule
\multicolumn{9}{c}{\emph{Registration Recall} $\uparrow$} \\
\midrule
Baseline & 0.611 & 0.444 & 0.301 & \textbf{0.205} & 0.502 & 0.455 & 0.499 & 0.431 \\
Sim(3)-FastMAC  & 0.444 & 0.445 & 0.400 & 0.0 & 0.577 & 0.091 & 0.501 & 0.351 \\
Sim(3)-SC2-PCR  & 0.446 & 0.222 & 0.195 & 0.103 & 0.385 & 0.088 & 0.378 & 0.260 \\
Sim(3)-SC2-PCR++  & 0.172 & 0.155 & 0.0 & 0.099 & 0.308 & 0.0 & 0.125 & 0.123 \\
MAC      & N/A & N/A & N/A & N/A & N/A & N/A & N/A & N/A \\
Sim(3)-TurboReg & 0.278 & 0.281 & 0.304 & 0.101 & 0.462 & 0.455 & 0.250 & 0.304 \\
\rowcolor{gray!20} Ex-Sim(3)-Reg     & \textbf{0.833} & \textbf{0.722} & \textbf{0.600} & {0.186} & \textbf{0.808} & \textbf{0.545} & \textbf{0.750} & \textbf{0.635} (\textbf{+20.4\%})\\
\bottomrule
\end{tabular}}
\vspace{-6pt}
\label{table:cross_res_2}
\end{table}

\begin{table}[t]
\setlength{\tabcolsep}{2pt}
\centering
\caption{
Correspondence pruning performance on the RGBD-v2 dataset~\cite{RGBD-dataset} under low-inlier conditions. 
We restrict the inlier number to simulate extreme outlier scenarios. These highlight robustness when geometric constraints are limited.
}
\begin{tabular}{l|ccccc}
\toprule
Method  & RG-11 & RG-12 & RG-13 & RG-14 & Mean \\
\midrule
\multicolumn{6}{c}{\emph{Inlier Ratio} $\uparrow$} \\
\midrule
Baseline & 0.195 & 0.271 & 0.075 & 0.123 & 0.166 \\
Sim(3)-FastMAC  & 0.241 & 0.318 & 0.085 & 0.138 & 0.195 \\
Sim(3)-SC2-PCR  & 0.334 & 0.392 & 0.112 & 0.159 & 0.249 \\
Sim(3)-SC2-PCR++  & 0.231 & 0.294 & 0.085 & 0.130 & 0.185 \\
Sim(3)-MAC      & 0.244 & 0.311 & 0.093 & 0.135 & 0.196 \\
Sim(3)-TurboReg & 0.239 & 0.309 & 0.091 & 0.134 & 0.193 \\
\rowcolor{gray!20} Ex-Sim(3)-Reg     & \textbf{0.444} & \textbf{0.514} & \textbf{0.144} & \textbf{0.206} & \textbf{0.327} (\textbf{+16.1\%}) \\
\midrule
\multicolumn{6}{c}{\emph{Registration Recall} $\uparrow$} \\
\midrule
Baseline & 0.347 & 0.443 & 0.077 & 0.314 & 0.295 \\
Sim(3)-FastMAC  & 0.431 & 0.533 & 0.128 & 0.236 & 0.332 \\
Sim(3)-SC2-PCR  & 0.556 & 0.484 & 0.171 & 0.210 & 0.355 \\
Sim(3)-SC2-PCR++  & 0.153 & 0.410 & 0.111 & 0.144 & 0.204 \\
Sim(3)-MAC      & 0.292 & 0.607 & 0.145 & 0.295 & 0.335 \\
Sim(3)-TurboReg & 0.486 & 0.484 & 0.103 & 0.303 & 0.344 \\
\rowcolor{gray!20} Ex-Sim(3)-Reg     & \textbf{0.708} & \textbf{0.852} & \textbf{0.231} & \textbf{0.376} & \textbf{0.542} (\textbf{+24.7\%}) \\
\bottomrule
\end{tabular}
\vspace{-5pt}
\label{table:few_res_1}
\end{table}

\begin{table}[t]
\setlength{\tabcolsep}{2pt}
\centering
\caption{
Cross-domain I2P registration performance on the ScanNet dataset~\cite{scan-net}. We integrate the proposed Ex-Sim(3)-Reg (Ours) into different baselines, which demonstrates the superior improvements.
}
\resizebox{1.0\linewidth}{!}{
\begin{tabular}{c|c>{\columncolor{gray!20}}c|c>{\columncolor{gray!20}}c|c>{\columncolor{gray!20}}c|c>{\columncolor{gray!20}}c|c>{\columncolor{gray!20}}c}
\toprule
Method & Matr & +Ours & FreeReg & +Ours & Top-I2P & +Ours & MinCD & +Ours & Diff-I2P & +Ours  \\
\hline
IR & 0.451 & \textbf{0.662}  & 0.568 & \textbf{0.690} & 0.493 & \textbf{0.674} & 0.464 & \textbf{0.678} & 0.521 & \textbf{0.682} \\
RR & 0.842 & \textbf{0.943} & 0.780 & \textbf{0.932} & 0.901 & \textbf{0.950} & 0.898 & \textbf{0.957} & 0.914 & \textbf{0.963}  \\
\bottomrule
\end{tabular}}
\vspace{-5pt}
\label{table:exp_new_added}
\end{table}

\begin{figure*}[t]
	\centering
		\includegraphics[width=1\linewidth]{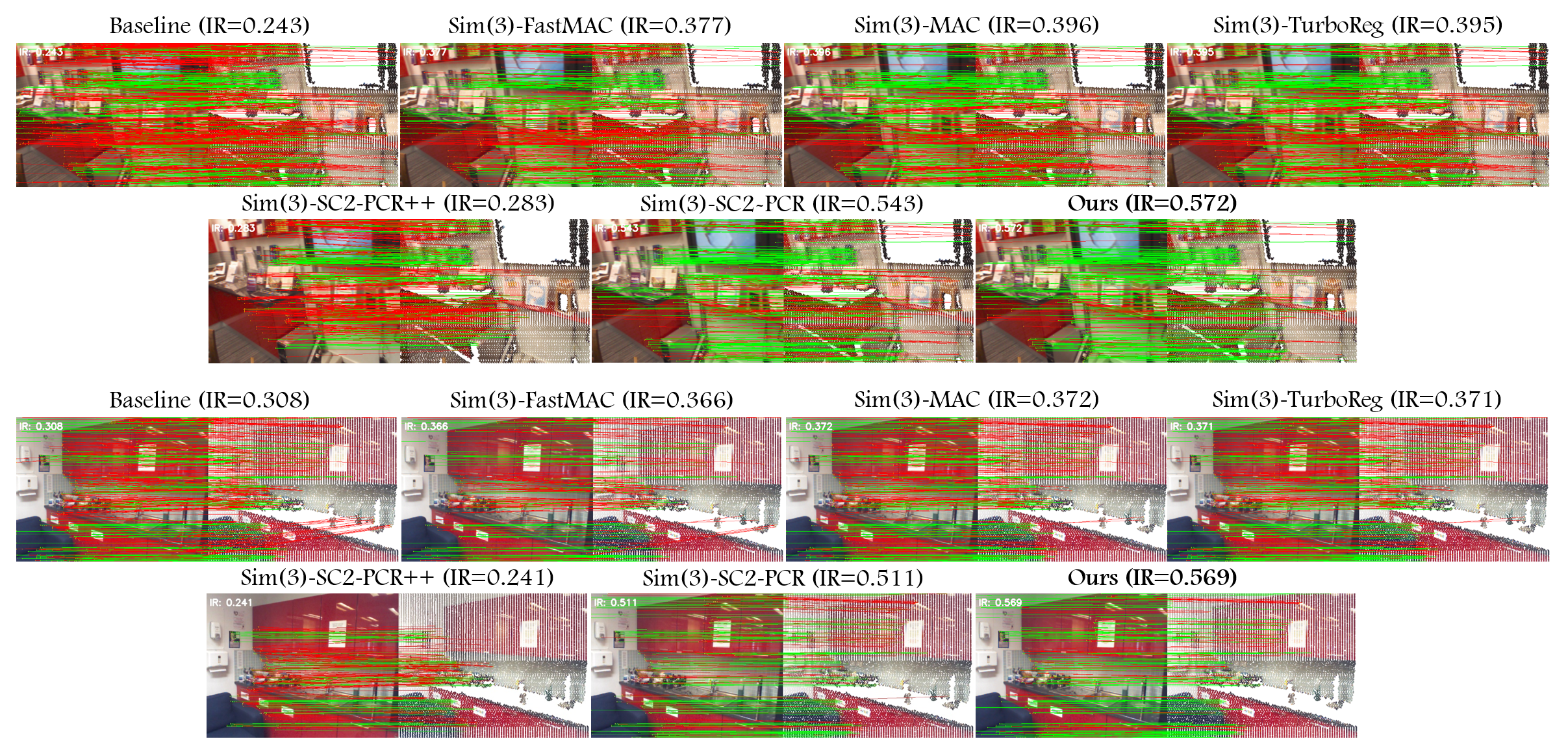}	
		\caption{Qualitative comparison on challenging scenes. Competing methods either remove excessive inliers or retain geometrically inconsistent matches. Our method preserves denser and spatially coherent inlier sets.}
	\label{fig:vis_main}
    \vspace{-6pt}
\end{figure*}

\vspace{+1mm}
\noindent\textbf{Performances under unseen condition}. 
We further assess cross-domain generalization by training on one dataset and testing on unseen target datasets. In the 7-Scenes $\rightarrow$ RGBD-v2 and 7-Scenes $\rightarrow$ ScanNet experiments, our method achieves RR gains of \textbf{13.1\%} and \textbf{20.4\%}, respectively (Tables~\ref{table:cross_res_1} and~\ref{table:cross_res_2}). The larger improvement on ScanNet reflects its greater scene diversity and varying depth-scale distortions, which pose greater challenges to existing methods. Standard Sim (3)-based pruning methods fail to outperform the baseline, because they overlook the noises in predicted depths, overfitting to false-positive inliers. In contrast, our extended Sim(3) based method explicitly compensates for noisy depth priors, yielding stable I2P registration performances.
Additional metrics of the pose errors are reported in the supplemental materials.   

\vspace{+1mm}
\noindent\textbf{Performances under low-inlier condition}. We evaluate the pruning performance under the condition of extremely few inliers. A low-inlier correspondence dataset is generated from the RGBD-v2 dataset \cite{RGBD-dataset}, containing sparse point clouds and low-quality RGB images. This setting is created by inferring I2P correspondences from a model trained with insufficient supervision. In such a condition, the traditional RANSAC-based P3P and regular Sim(3)-based pruning methods typically fail due to insufficient geometric constraints. As presented in Table~\ref{table:few_res_1}, our method achieves the highest IR and RR, exceeding 50\% RR even under severe correspondence sparsity. This is because Ex-Sim(3)-Reg leverages a family of extended compatibility graphs to reliably discover potential inliers, thus enabling robust registration. 

\vspace{+1mm}
\noindent\textbf{Qualitative comparison}.
Figure~\ref{fig:vis_main} visualizes correspondence pruning results. With the extended Sim(3) registration model, our method recovers denser and geometrically consistent inlier sets. 


\vspace{+1mm}
\noindent\textbf{Comparison to I2P registration}. Moreover, we integrate the proposed pruning method into the state-of-the-art I2P registration methods, such as FreeReg \cite{freereg}, Top-I2P \cite{top-i2p}, and MinCD \cite{mincd}. Methods except FreeReg (a zero-shot method) are all trained on 7-Scenes \cite{scene-7-dataset} but tested on ScanNet \cite{scan-net}. From Table \ref{table:exp_new_added}, it is observed that Ex-Sim(3)-Reg improves the performance of current methods, which validates the practical value of our pruning pipeline. 

\vspace{+1mm}
\noindent\textbf{Results analysis}. Several patterns can be observed from the above results. Standard Sim(3)-based pruning ignores depth prediction errors, leading to only marginal or even negative RR gains. Sim(3)-SC2-PCR++~\cite{sc2-pcr++} uses a strict inlier criterion designed for 3D registration,
but fails to account for point cloud distortion from noisy depth maps, resulting in unstable I2P registration. Sim(3)-TurboReg~\cite{turboreg} focuses on identifying the optimal cliques with minimal elements, and thus exhibits robustness to the noisy depths than others based on regular Sim(3) registration. Different from the previous methods, the proposed extended Sim(3) based pruning method naturally models depth noise, explicitly corrects noisy correspondences, thus achieving the most consistent and robust performance across all settings.

\begin{table}[t]
\setlength{\tabcolsep}{2pt}
\centering
\caption{
Effect of the number of sub-intervals $K$ on IR, RR, and runtime. Increasing $K$ improves accuracy but with increased computational overhead.
}
\begin{tabular}{c|cc>{\columncolor{gray!20}}ccc}
\toprule
$K$ & 4 & 6 & 8 & 10 & 12  \\
\hline
IR/RR & 0.505/0.362 & 0.542/0.394 & 0.548/\textbf{0.472} & 0.573/0.448 & \textbf{0.682}/0.427 \\
Time & \textbf{57.6ms} & 65.4ms & 73.6ms & 81.7ms & 89.5ms \\
\bottomrule
\end{tabular}
\vspace{-5pt}
\label{table:ab_1}
\end{table}

\begin{table}[t]
\setlength{\tabcolsep}{2pt}
\centering
\caption{
Impact of pixel reprojection threshold $\delta_{thr}$ on IR, RR, and relative translation error (RTE). Larger thresholds improve robustness but may admit additional outliers, revealing the robustness–precision trade-off.
}
\begin{tabular}{c|cc>{\columncolor{gray!20}}ccc}
\toprule
$\delta_{thr}$ & 8 & 12 & 16 & 20 & 24  \\
\hline
IR/RR & 0.513/0.380 & 0.535/0.466 & \textbf{0.548/0.472} & 0.521/0.384 & 0.505/0.372 \\
RTE & 0.072m & 0.068m & \textbf{0.067m} & 0.070m & 0.071m \\
\bottomrule
\end{tabular}
\vspace{-5pt}
\label{table:ab_2}
\end{table}

\begin{table}[t]
\setlength{\tabcolsep}{2pt}
\centering
\caption{
Ablation studies on MAC search and scale-correction function. Removing the MAC search significantly degrades inlier consistency, while omitting the correction function reduces robustness to depth noise. Runtime comparison illustrates the computational trade-offs.
}
\begin{tabular}{cc|c|cc}
\toprule
MAC~\cite{mac} & FastMAC~\cite{fastmac} & Scale-correction function $\varphi(\cdot)$ &  RR & Average Runtime/ms  \\
\hline
& $\checkmark$ &  & 0.311 & \textbf{72.4} \\
$\checkmark$ &  & $\checkmark$ & \textbf{0.485} & 204.8 \\
\rowcolor{gray!20} & $\checkmark$ & $\checkmark$  & 0.472 & 73.6 \\
\bottomrule
\end{tabular}
\vspace{-5pt}
\label{table:ab_3}
\vspace{-6pt}
\end{table}

\subsection{Ablation Studies} 
\label{sec.exp.C}

We investigate the effect of key hyperparameters and components of the proposed method on the noisy 7-Scenes dataset \cite{scene-7-dataset}.

\vspace{+1mm}
\noindent\textbf{Number of sub-intervals $K$}. Increasing $K$ can enlarge the upper bound of inliers, since the size of $\cup_{k=1}^K \mathscr{F}_k$ is growing with $K$. As shown in Table~\ref{table:ab_1}, performance saturates when $K > 6$, while the computational cost continues to grow. We thus adopt $K=6$ to balance accuracy and efficiency. 

\vspace{+1mm}
\noindent\textbf{Pixel error threshold $\delta_{thr}$}. This parameter can weigh up the robustness and accuracy of Ex-Sim(3)-Reg. A smaller $\delta_{thr}$ benefits the precise poses but degrades robustness when the initial pose $\tilde{\mathbf{R}}$, $\tilde{\mathbf{t}}$ are noisy. Table~\ref{table:ab_2} shows that $\delta_{thr}=16$ provides the best trade-off between IR, RR, and RTE metrics, and is therefore used in our experiments. 


\vspace{+1mm}
\noindent\textbf{MAC search and  scale-correction}. We further investigate the effects of using FastMAC~\cite{fastmac} versus MAC~\cite{mac}, and scale-correction function.
As summarized in Table~\ref{table:ab_3}, MAC achieves slightly higher RR but is $2.78\times$ slower. FastMAC offers a favorable accuracy-efficiency trade-off.
Correction function substantially improves accuracy across both variants, confirming its necessity for reliable correspondence pruning.


\section{Conclusions}
\label{sec.conc}

In this paper, we reformulate depth prior-based 2D-3D correspondence pruning as an extended Sim(3) registration problem. Different from regular Sim(3), the extended Sim(3) can account for errors in depth priors, thus enabling the robust pruning performance. However, extended Sim(3) registration is an under-determined problem, and no previous work has specifically studied it. To address this problem, we propose a heuristic yet effective pruning algorithm named Ex-Sim(3)-Reg. By detecting and correcting maximum cliques from a family of extended compatibility graphs, Ex-Sim(3)-Reg simplifies extended Sim(3) to the standard SE(3) registration with a high-inlier set, allowing accurate and stable camera pose estimation. Moreover, theoretical analysis is conducted to justify the effectiveness of Ex-Sim(3)-Reg. Finally, experiments on four public datasets verify that our approach significantly outperforms state-of-the-art methods.

\vspace{+2mm}
\noindent\textbf{Limitation and future work.} Despite the superior performances, our method can recover at most the inlier set $\cup_{k=1}^K \mathscr{F}_k$. Even when $K$ is $+\infty$, $\cup_{k=1}^{+\infty} \mathscr{F}_k$ remains a subset of the true inlier set. To search for more potential inliers, we will refine the construction and usage of extended compatibility graphs in future work. 

\vspace{+1mm}
\noindent\textbf{Acknowledgments}. The authors greatly thank Prof. You Yang for his suggestions. This work is partially supported by the National Natural Science Foundation of China (Grand ID: 62502171, 62372377) and China Postdoctoral Science Foundation (Grand ID: 2024M761014, GZC20252285)




\appendix

\section{More Theoretical Discussions} 

\begin{figure*}[h]
	\centering
		\includegraphics[width=0.8\linewidth]{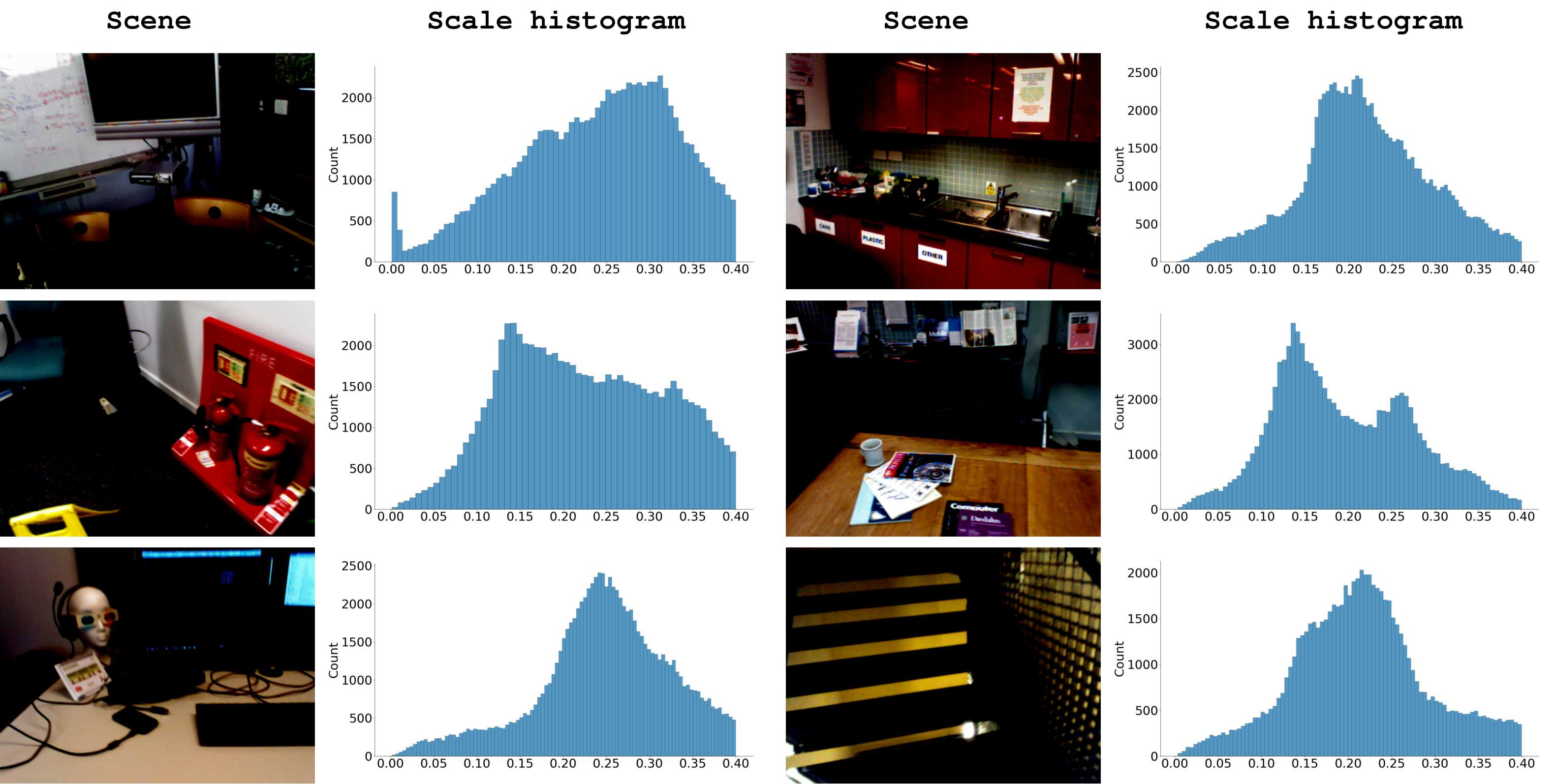}	
		\caption{Histogram of scales under different scale values in the 7-Scenes dataset~\cite{scene-7-dataset}, illustrating the prevalence of $s_i$ inconsistencies in real applications, which supports the use of the extended Sim(3).}
	\label{fig:supp_scales}
    \vspace{-12pt}
\end{figure*}

\vspace{+2mm}
\noindent\textbf{Necessity of extended Sim(3)}. In the main paper, we use the extended Sim(3) transformation instead of the regular Sim(3) transformation. To support this, we conduct a statistical verification on the 7-Scenes dataset \cite{scene-7-dataset}. Using both the ground-truth and predicted depths, we can compute the depth error $\delta_i$ of each inlier and then visualize the distribution of $s_i=s\cdot\delta_i$ in Fig.~\ref{fig:supp_scales}. The results show that $s_i$ follows a long-tailed distribution in complex scenes. This indicates that errors in depth priors cannot be neglected in the correspondence pruning.

\vspace{+2mm}
\noindent\textbf{Details of proposed algorithm}. We add more details of the proposed algorithm in a question-answering manner: 

\noindent {\textit{$(1)$ Why adaptive voting is reasonable?}} Because distribution of $s$ (computed from inliers) is {concentrated in a fixed and small interval (Fig. \ref{fig:supp_scales})}. This observation enables us to use adaptive voting in work \cite{TEASER} to estimate the average scale $\tilde{s}$.

\noindent {\textit{$(2)$ Why extra search needed around $\tilde{s}$?}} Because the problem of Ex-Sim(3)-Reg in Eq. 4 needs to estimate $s_i$ of all inliers. As $s_i$ is near at $\tilde{s}$, we design an extra search on $[\tilde{s}-\Delta s,\tilde{s}+\Delta s]$ to determine both $s_i$ and inliers. 

\noindent \textit{$(3)$ Selection of $\Delta s$?} Fig. \ref{fig:supp_scales} shows that $s$ lies in a interval of $[0,0.4]$ with the middle of $\tilde{s}$. The search on $[\tilde{s}-\Delta s,\tilde{s}+\Delta s]$ ($\Delta s =0.1\tilde{s}\sim 0.3\tilde{s}$) can cover most inliers. So, we empirically set $\Delta s=0.15\tilde{s}$.

\noindent \textit{$(4)$ Theory explanation.} Based on \textbf{inlier set maximization assumption}, inliers can be found in the maximum clique (Mac) of the compatibility graph built on 3D-3D correspondences \cite{mac}. If $\mathbf{d}_i\triangleq(\mathbf{q}_i, \mathbf{p}_i) \in \mathrm{Mac}(\mathcal{G}_{\textrm{SOG}}(\zeta))$, $\{\mathbf{d}_i\}$ is an inlier candidate set satisfying $\mathbf{p}_i=\mathbf{R}(\zeta\mathbf{q}_i)+\mathbf{t}$.  It means that {every hypothesis $\zeta$ can generate a standard 3D-3D inlier candidate set $\{\mathbf{e}_i\}$ where $\mathbf{e}_i\triangleq(\zeta\mathbf{q}_i, \mathbf{p}_i)$}. After that, we can collect a large inlier candidate set $\{\mathbf{e}_i\}$ by merging inlier sets from all hypotheses $\zeta$ in $(0,+\infty)$. In this way, Ex-Sim(3)-Reg is simplified as an SE(3) registration task on $\{\mathbf{e}_i\}$, solved by Eq. 7 with off-the-shelf 3D pruning methods. 

\begin{figure}[t]
	\centering
		\includegraphics[width=1.0\linewidth]{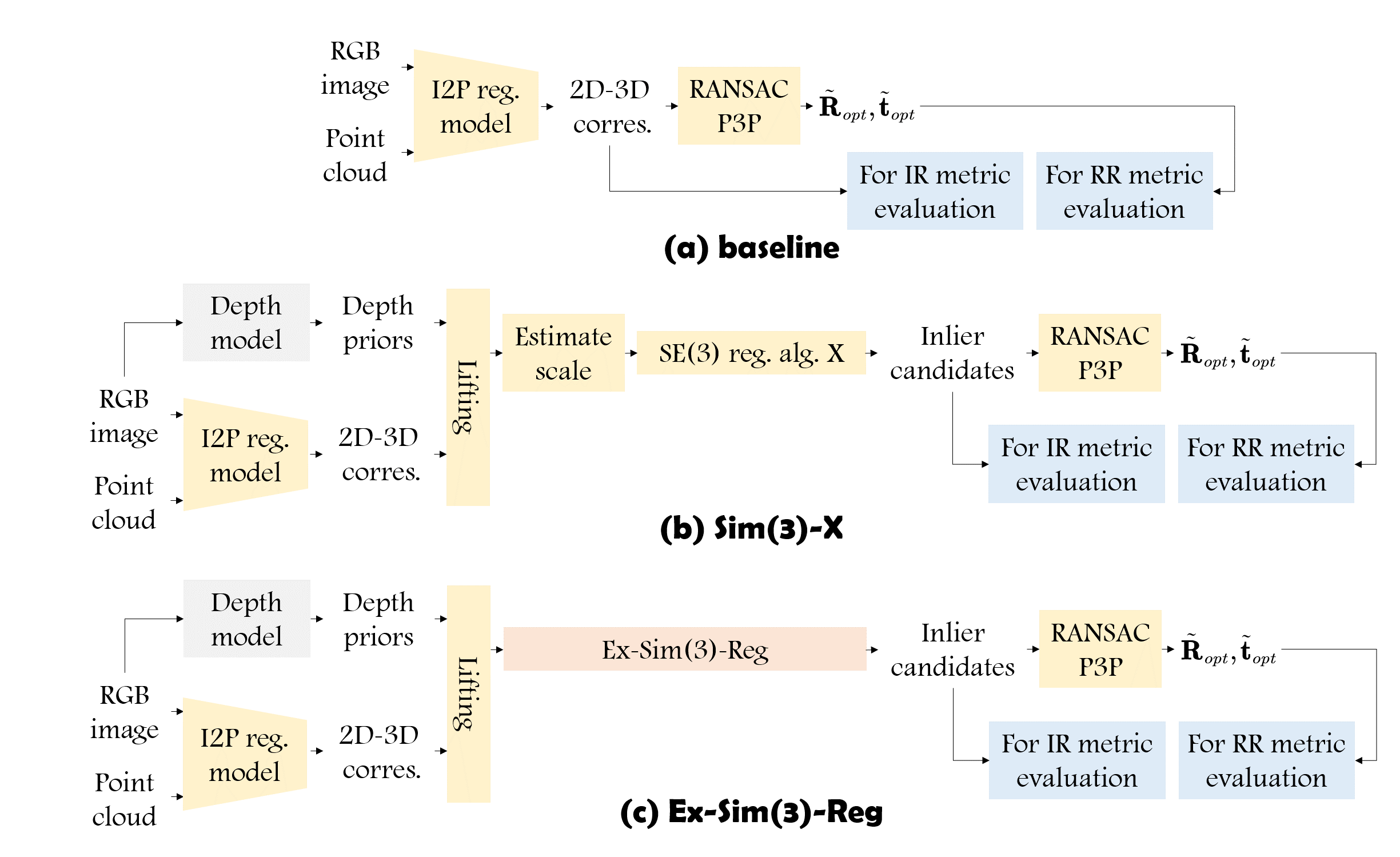}	
		\caption{Evaluation protocol. Inlier Ratio (IR) measures the quality of inlier candidates, while Registration Recall (RR) evaluates the accuracy of the estimated rotation $\mathbf{\tilde{R}}_{opt}$ and translation $\mathbf{\tilde{t}}_{opt}$. (a) \textbf{Baseline} denotes the case without pruning. (b) \textbf{Sim(3)-X} first estimates a global scale $\tilde{s}$~\cite{TEASER}, converts correspondences from $\{\mathbf{d}_i\}_{i=1}^\mathrm{N}$ to $\{(\tilde{s}\mathbf{q}_i, \mathbf{p}_i)\}_{i=1}^\mathrm{N}$, and then applies an SE(3) solver \textbf{X} to identify inliers. (c) Ex-Sim(3)-reg is the proposed pruning method. Iteration time in RANSAC P3P is the same (1K) in all compared methods.}
	\label{fig:eval_full}
    \vspace{-6pt}
\end{figure}

\begin{figure*}[t]
	\centering
		\includegraphics[width=1\linewidth]{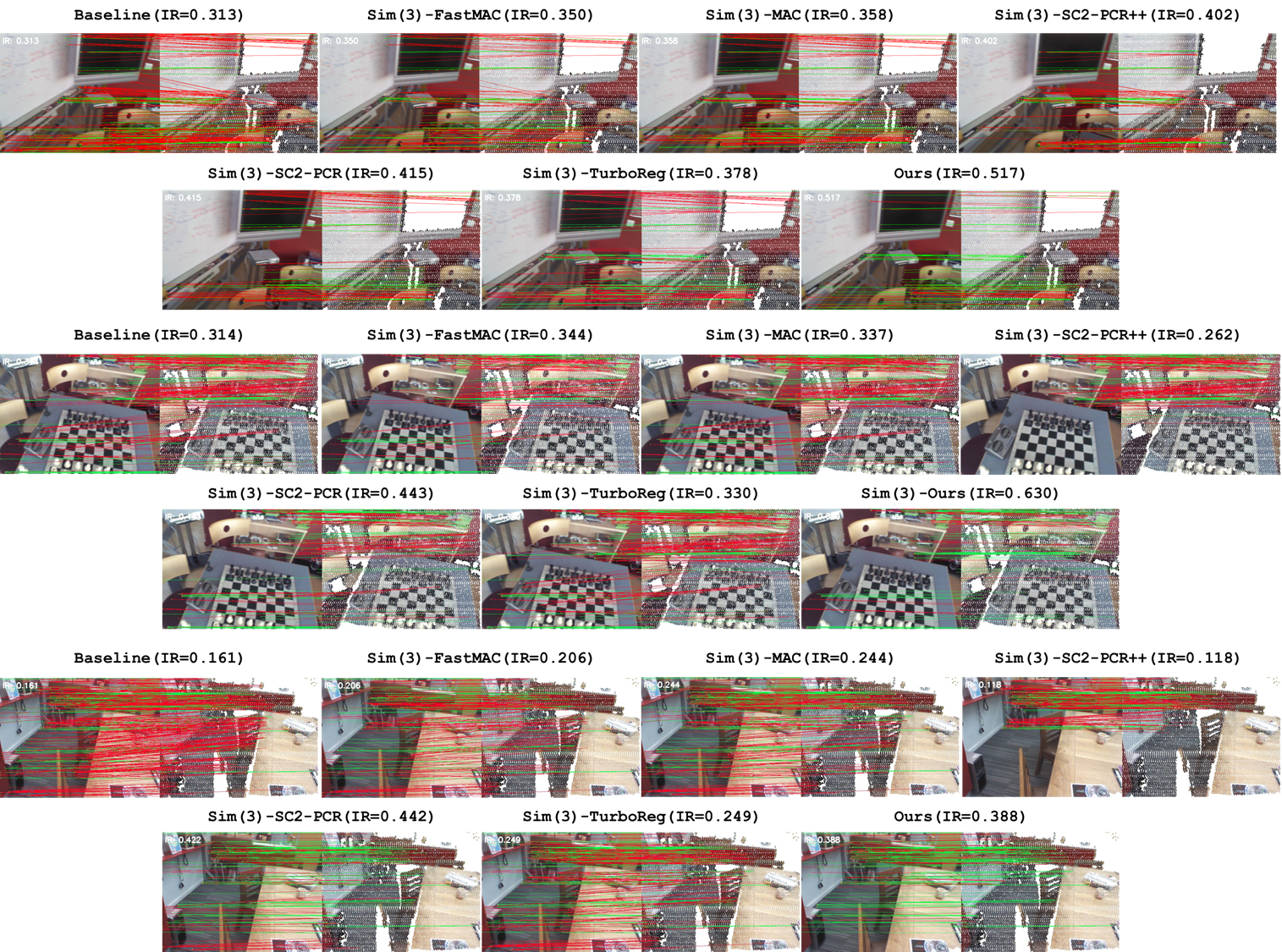}	
		\caption{Qualitative comparisons of different methods on the 7-Scenes dataset~\cite{scene-7-dataset}. \textcolor{green}{\textbf{Green}} and \textcolor{Red}{\textbf{red}} lines denote correct and incorrect correspondences, respectively.}
	\label{fig:supp-vis-1}
    \vspace{-12pt}
\end{figure*}

\vspace{+2mm}
\noindent\textbf{Complexity analysis}. The overall computational complexity is determined by three core steps: (i) scale hypothesis sampling, (ii) inlier candidate search, and (iii) pose estimation. 

\begin{table}[t]
\setlength{\tabcolsep}{2pt}
\centering
\caption{
Key hyperparameters of the proposed Ex-Sim(3)-Reg method.
}
\resizebox{0.50\linewidth}{!}{
\begin{tabular}{c|c}
\toprule
Hyperparameters & Value \\
\midrule
Number of intervals ($K$) & 8 \\
Threshold for 3D distance error ($d_{thr}$) & 0.2 \\
Threshold for 2D reprojection error ($\delta_{thr}$) & 16 \\
Scale interval ($\Delta s$) & $0.15 \tilde{s}$ \\ 
\bottomrule
\end{tabular}}
\label{table:params}
\end{table}

\begin{table}[!ht]
\setlength{\tabcolsep}{2pt}
\centering
\caption{
Experimental settings for correspondence pruning. $N_{\text{corr}}$ denotes the average number of 2D-3D correspondences, and IR denotes the average inlier ratio before pruning. These statistics reflect the difficulty level of each dataset.
}
\resizebox{0.85\linewidth}{!}{
\begin{tabular}{c|ccc}
\toprule
Pruning settings & Implementation scheme & $N_{\text{corr}}$ & IR  \\
\midrule
Disturbed condition & Adding noise on 3D point cloud & Nearly 0.1K & Nearly 25\% \\
Unseen condition & Open-domain testing & Nearly 1.0K & Nearly 30\% \\
Few-inlier condition & Training with insufficient samples & Nearly 0.1K & Nearly 15\% \\
\bottomrule
\end{tabular}}
\vspace{-5pt}
\label{table:data_settings}
\end{table}

\begin{figure*}[t]
	\centering
		\includegraphics[width=1\linewidth]{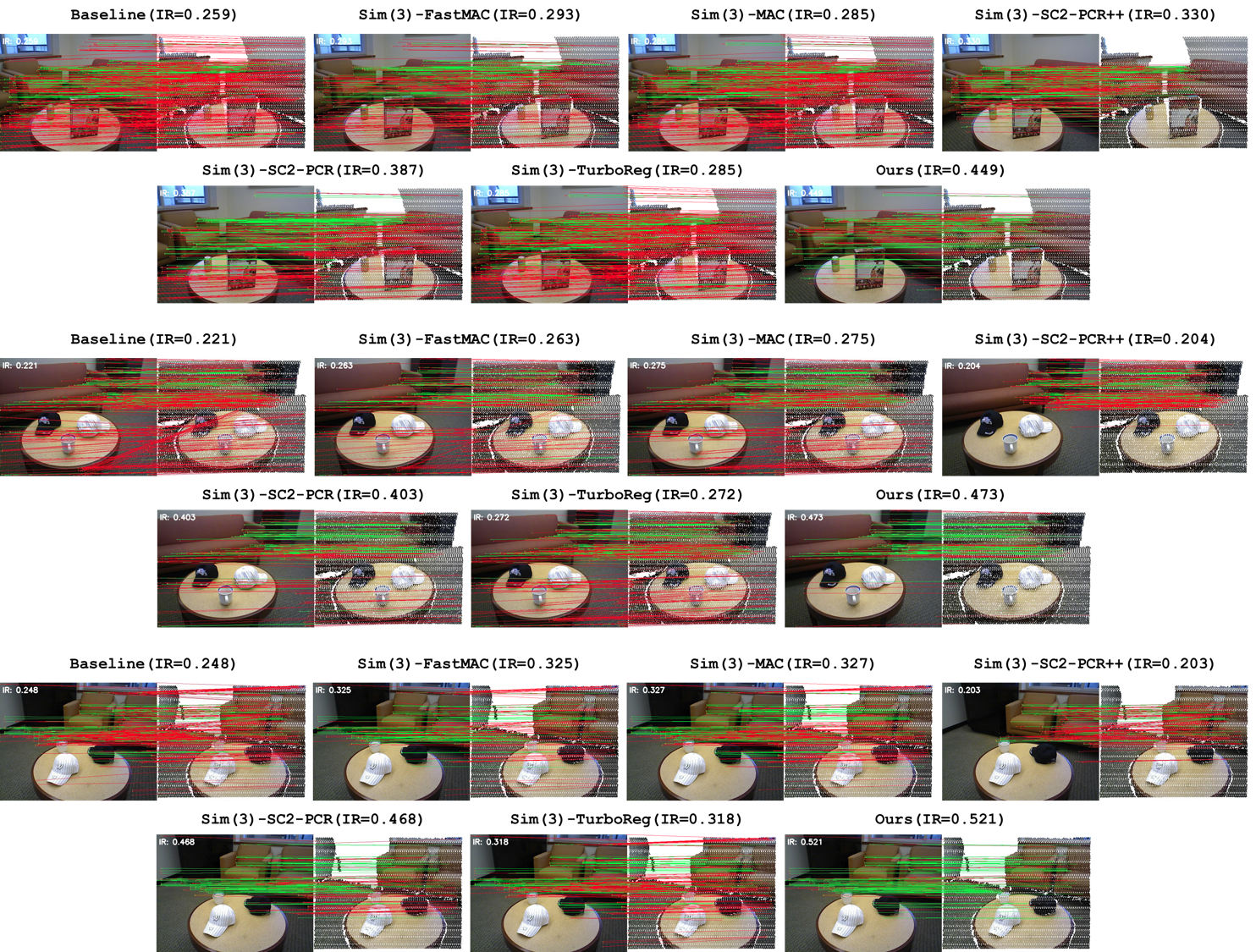}	
		\caption{Qualitative comparisons of different methods on the RGBD-v2 dataset~\cite{RGBD-dataset}.  \textcolor{green}{\textbf{Green}} and \textcolor{red}{\textbf{red}} lines indicate correct and incorrect correspondences, respectively.}
	\label{fig:supp-vis-2}
\end{figure*}

\begin{figure*}[t]
	\centering
		\includegraphics[width=1\linewidth]{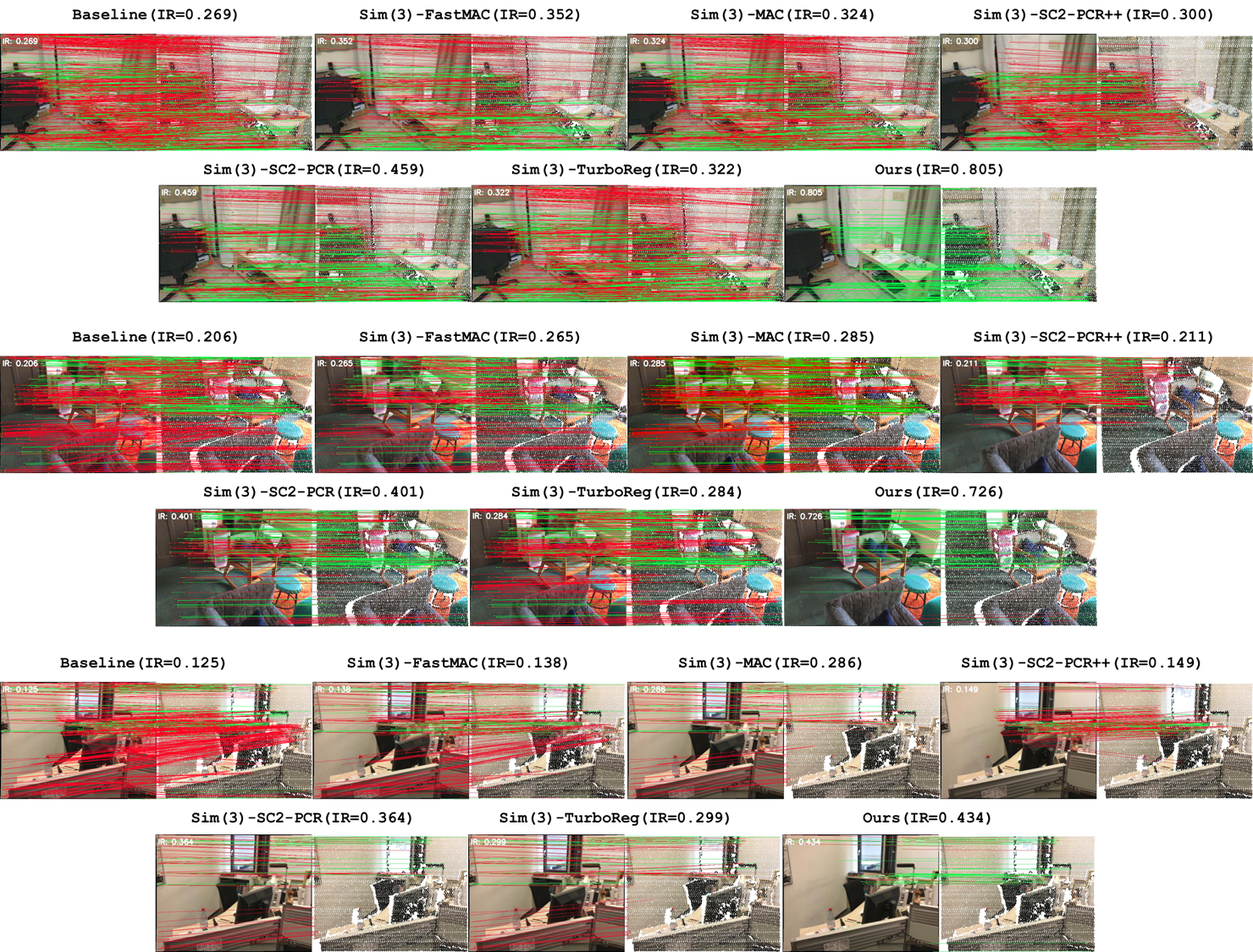}	
		\caption{Qualitative comparisons of different methods on the ScanNet dataset~\cite{scan-net}. \textcolor{green}{\textbf{Green}} and \textcolor{Red}{\textbf{red}} lines indicate correct and incorrect correspondences, respectively.}
	\label{fig:supp-vis-3}
\end{figure*}

\begin{figure*}[!ht]
	\centering
		\includegraphics[width=1\linewidth]{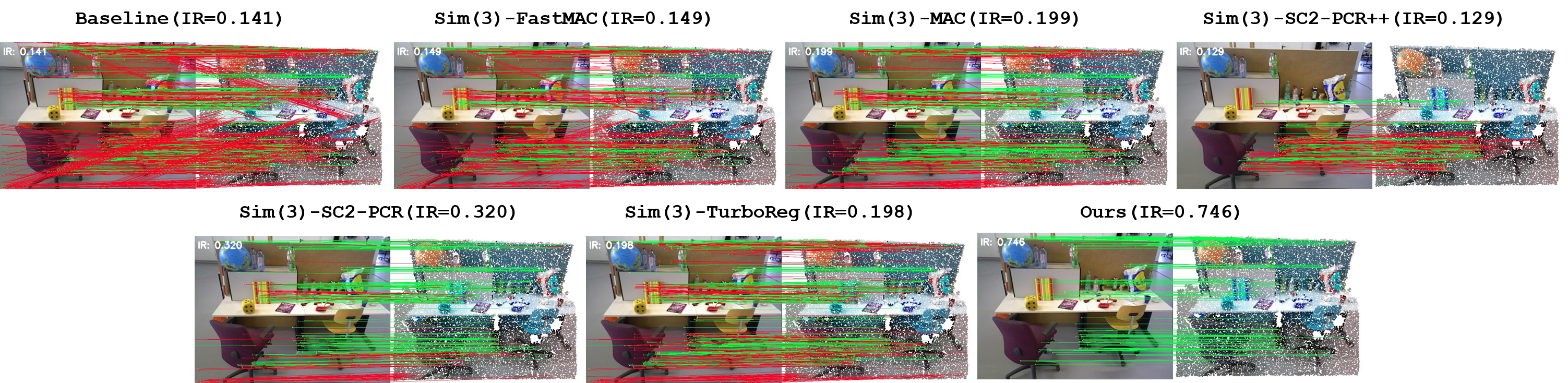}	
		\caption{Qualitative comparisons of different methods on the TUM dataset~\cite{tum}.  \textcolor{green}{\textbf{Green}} and \textcolor{Red}{\textbf{red}} lines indicate correct and incorrect correspondences, respectively.}
	\label{fig:supp-vis-4}
\end{figure*}

\begin{figure*}[!ht]
	\centering
		\includegraphics[width=1\linewidth]{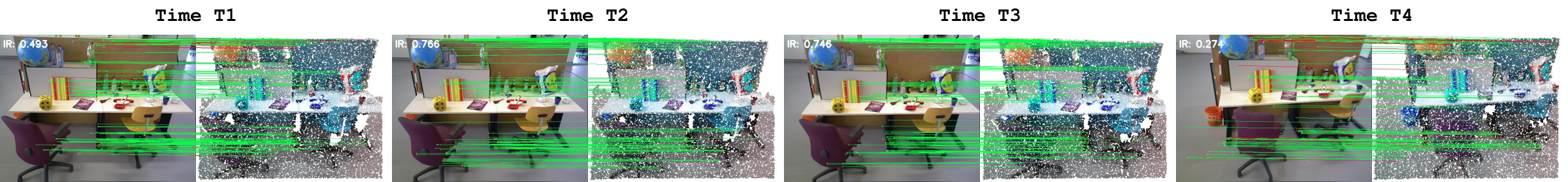}	
		\caption{Qualitative comparisons of the proposed method at different timestamps on the TUM dataset~\cite{tum}. }
	\label{fig:supp-vis-5}
\end{figure*}

\begin{itemize}
    \item Step (i) constructs a histogram with complexity $\mathcal{O}(Bin)$, where $Bin$ denotes the number of bins \cite{TEASER}. Using the API of \texttt{Torch.histogram}, this step takes only a few milliseconds. 

    \item Step (ii) has complexity $\mathcal{O}(K)\cdot\mathcal{O}(\mathrm{MAC\_Search})$. We adopt FastMAC~\cite{fastmac} with stochastic spectral sampling, which reduces the standard MAC~\cite{mac} search complexity from $\mathcal{O}(N^3)$ to $\mathcal{O}(N^2)$ for $N$ correspondences, resulting in an overall complexity of $\mathcal{O}(KN^2)$. In our experiments, FastMAC requires only 4-10 milliseconds. 

    \item Step (iii) uses TurboReg (ICCV'25)~\cite{turboreg} and RANSAC-based P3P~\cite{ransac-p3p} for 3D correspondence pruning, taking only 10-20 milliseconds. 
\end{itemize}

\section{More Experimental Details and Analysis}

\vspace{+2mm}
\noindent\textbf{Implementation details}. In the first step of Ex-Sim(3)-Reg, we define an interval $(\tilde{s}-\Delta s, \tilde{s}+\Delta s)$ for sampling $\zeta$. $\Delta s$ is set to $0.15 \tilde{s}$. In the second and third steps, FastMAC~\cite{fastmac} and TurboReg~\cite{turboreg} are integrated into our pipeline, and all their hyperparameters follow the default indoor configuration (as our experiments are primarily conducted on indoor datasets). The threshold $d_{thr}$ is set following prior work~\cite{sc2-pcr}. For clarity, all hyperparameters are summarized in Table~\ref{table:params}. Additionally, the differences between the three experiment settings are detailed in Table~\ref{table:data_settings}. The complete evaluation protocol is presented in Fig. \ref{fig:eval_full}. In the metric evaluation, the threshold of the RR metric is 5cm for 7-Scenes \cite{scene-7-dataset} and TUM \cite{tum}, 2.5cm for ScanNet \cite{scan-net}, and 10cm for the RGBD-v2 dataset \cite{RGBD-dataset}. 


\begin{table}[t]
\setlength{\tabcolsep}{2pt}
\centering
\caption{
Evaluation on the noisy 7-Scenes dataset.
\textbf{Bold} numbers denote the best mean performance in terms of RRE.
}
\resizebox{1.0\linewidth}{!}{
\begin{tabular}{c|ccccc>{\columncolor{gray!20}}c}
\toprule
Method & Sim(3)-FastMAC & Sim(3)-SC2-PCR & Sim(3)-SC2-PCR++ & Sim(3)-MAC & Sim(3)-TurboReg & Ours \\
\midrule
RTE/m & \textbf{0.066} & 0.067 & 0.085 & 0.075 & 0.068 & \textbf{0.066} \\
RRE/deg & 2.426 & 2.375 & 3.005 & 2.539 & 2.381 & \textbf{2.367} \\
\bottomrule
\end{tabular}}
\vspace{-5pt}
\label{table:exp_1}
\end{table}

\begin{table}[t]
\setlength{\tabcolsep}{2pt}
\centering
\caption{
Evaluation on the noisy TUM dataset.
\textbf{Bold} numbers denote the best mean performance in terms of RRE.
}
\resizebox{1.0\linewidth}{!}{
\begin{tabular}{c|ccccc>{\columncolor{gray!20}}c}
\toprule
Method & Sim(3)-FastMAC & Sim(3)-SC2-PCR & Sim(3)-SC2-PCR++ & Sim(3)-MAC &  Sim(3)-TurboReg & Ours \\
\midrule
RTE/m & 0.059 & 0.051 & 0.057 & 0.064 & 0.058 & \textbf{0.036} \\
RRE/deg & 2.824 & 2.003 & 2.949 & 3.074 & 2.994 & \textbf{1.824} \\
\bottomrule
\end{tabular}}
\vspace{-5pt}
\label{table:exp_2}
\end{table}

\begin{table}[t]
\setlength{\tabcolsep}{2pt}
\centering
\caption{
Evaluation under an unseen testing setup (training on 7-Scenes, testing on ScanNet). 
\textbf{Bold} numbers denote the best mean performance in terms of RRE.
}
\resizebox{1.0\linewidth}{!}{
\begin{tabular}{c|ccccc>{\columncolor{gray!20}}c}
\toprule
Method & Sim(3)-FastMAC & Sim(3)-SC2-PCR & Sim(3)-SC2-PCR++ & Sim(3)-MAC &  Sim(3)-TurboReg & Ours \\
\midrule
RTE/m & 0.043 & 0.034 & 0.050 & N/A & 0.034 & \textbf{0.025} \\
RRE/deg & 1.470 & 1.008 & 2.127 & N/A & 1.095 & \textbf{0.751} \\
\bottomrule
\end{tabular}}
\vspace{-5pt}
\label{table:exp_3}
\end{table}

\begin{table}[!ht]
\setlength{\tabcolsep}{2pt}
\centering
\caption{
Evaluation on the challenging RGBD-v2 dataset with a low inlier ratio. \textbf{Bold} numbers denote the best mean performance in terms of RRE.
}
\resizebox{1.0\linewidth}{!}{
\begin{tabular}{c|ccccc>{\columncolor{gray!20}}c}
\toprule
Method & Sim(3)-FastMAC & Sim(3)-SC2-PCR & Sim(3)-SC2-PCR++ & Sim(3)-MAC &  Sim(3)-TurboReg & Ours \\
\midrule
RTE/m & 0.097 & 0.106 & 0.107 & 0.104 & 0.104 & \textbf{0.093} \\
RRE/deg & 2.854 & 3.111 & 3.145 & 2.947 & 2.913 & \textbf{2.839} \\
\bottomrule
\end{tabular}}
\vspace{-5pt}
\label{table:exp_4}
\end{table}

\begin{table}[!ht]
\setlength{\tabcolsep}{2pt}
\centering
\caption{
Runtime comparison (in milliseconds) of I2P pruning methods on the 7-Scenes dataset.
}
\resizebox{1\linewidth}{!}{
\begin{tabular}{c|cccccc}
\toprule
Method  & Sim(3)-FastMAC & Sim(3)-SC2-PCR & Sim(3)-SC2-PCR++ & Sim(3)-MAC & Sim(3)-TurboReg & Ours \\
\hline
Time/ms    & 23.1 & 24.2 & 105.3 & 150.4  & \textbf{15.4} & 73.6 \\
\bottomrule
\end{tabular}}
\vspace{-5pt}
\label{table:runtime}
\end{table}

\begin{figure}[t]
	\centering
		\includegraphics[width=1\linewidth]{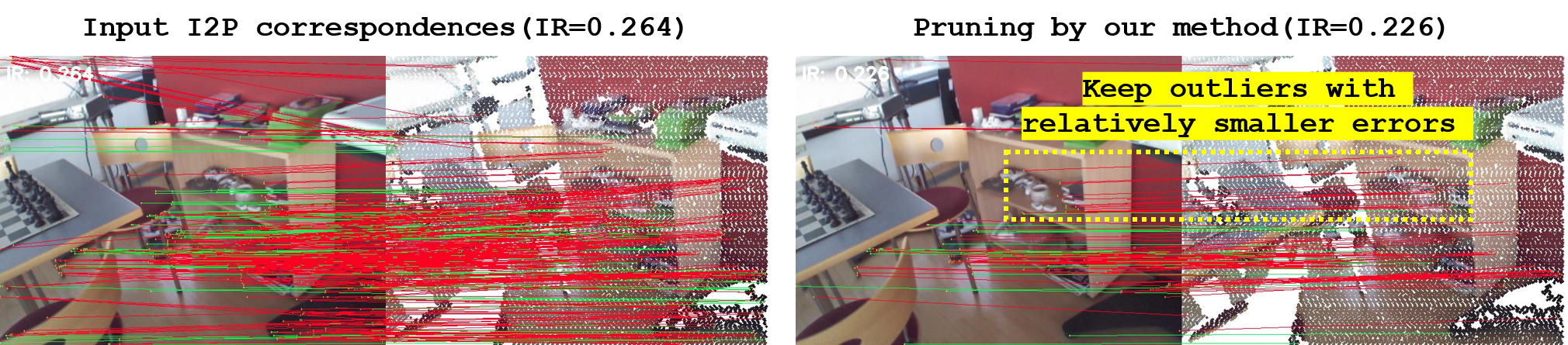}	
		\caption{Failure case of the proposed pruning method due to \textbf{unsuitable hyperparameter setting}. See Sec. A3 for analysis.  \textcolor{green}{\textbf{Green}} and \textcolor{red}{\textbf{red}} lines indicate the correct and incorrect correspondences, respectively.}
	\label{fig:supp-failure}
\end{figure}

\vspace{+2mm}
\noindent\textbf{Comparisons using RTE and RRE metrics}. In addition to IR and RR metrics, relative translation error (RTE) and relative rotation error (RRE) are also key metrics for evaluating registration accuracy~\cite{diff_reg_match}. We present the evaluations on the 7-Scenes~\cite{scene-7-dataset}, TUM~\cite{tum}, ScanNet~\cite{scan-net}, and RGBD-v2~\cite{RGBD-dataset} datasets in Tables~\ref{table:exp_1},~\ref{table:exp_2},~\ref{table:exp_3}, and~\ref{table:exp_4} where we compare our method with Sim(3)-FastMAC~\cite{fastmac}, Sim(3)-SC2-PCR~\cite{sc2-pcr}, Sim(3)-SC2-PCR++~\cite{sc2-pcr++}, Sim(3)-MAC~\cite{mac}, and Sim(3)-TurboReg~\cite{turboreg}. RTE is measured in meters, and RRE is measured in degrees. Our method achieves substantial improvements on both metrics, particularly on the TUM~\cite{tum} and ScanNet datasets~\cite{scan-net}, demonstrating its effectiveness in enhancing I2P registration accuracy. 

\begin{table*}[t]
\setlength{\tabcolsep}{2pt}
\centering
\caption{
Comparison of monocular depth estimation models. The baseline monocular depth estimation model used in our experiments is Depth Anything V2~\cite{dep_anything_v2}.
}
\resizebox{1.0\linewidth}{!}{
\begin{tabular}{c|cc|c|c|c}
\toprule
Model Names & Descriptions & Depth Accuracy & Runtime & Model Size & Used in Main Paper \\
\midrule
\texttt{depth\_anything\_v2\_metric\_hypersim\_vits} & \texttt{vits} & Low & nearly 32 ms & 99.2 MB & \\
\texttt{depth\_anything\_v2\_metric\_hypersim\_vitb} & \texttt{vitb} & Moderate & nearly 65 ms & 390.0 MB & \\
\texttt{depth\_anything\_v2\_metric\_hypersim\_vitl} & \texttt{vitl} & High & nearly 138 ms & 1.3 GB & $\checkmark$ \\
\bottomrule
\end{tabular}}
\vspace{-5pt}
\label{table:mde_models}
\end{table*}

\begin{table}[!ht]
\setlength{\tabcolsep}{2pt}
\centering
\caption{
Registration recall of different pruning methods using Depth Anything v2 with different model sizes on the TUM Freiburg\_01 scene~\cite{tum}.
}
\resizebox{0.8\linewidth}{!}{
\begin{tabular}{c|ccc}
\toprule
Models & Using \texttt{vits.pth} & Using \texttt{vitb.pth} & Using \texttt{vitl.pth} \\
\midrule
Sim(3)-TurboReg~\cite{turboreg} & 0.083 & 0.102 & 0.118 \\
Sim(3)-MAC~\cite{mac}           & 0.092 & 0.095 & 0.109 \\
Our method               & \textbf{0.333} & \textbf{0.317} & \textbf{0.277} \\
\bottomrule
\end{tabular}}
\vspace{-5pt}
\label{table:exp_mde_models}
\end{table}

\vspace{+2mm}
\noindent\textbf{Runtime comparisons}. Table~\ref{table:runtime} reports the runtime of Sim(3)-FastMAC~\cite{fastmac}, Sim(3)-SC2-PCR~\cite{sc2-pcr}, Sim(3)-SC2-PCR++~\cite{sc2-pcr++}, Sim(3)-MAC~\cite{mac}, Sim(3)-TurboReg~\cite{turboreg}, and our method. 
Due to the subgraph search procedure, our method is slower than Sim(3)-TurboReg~\cite{turboreg}, but it still outperforms Sim(3)-SC2-PCR++~\cite{sc2-pcr++} and Sim(3)-MAC~\cite{mac} in the accuracy-efficiency trade-off. In future work, we plan to integrate a parallel search module to improve computational efficiency.

\section{Qualitative Comparisons and Failure Analysis}

Qualitative results on the 7-Scenes~\cite{scene-7-dataset}, RGBD-v2~\cite{RGBD-dataset}, ScanNet~\cite{scan-net}, and TUM~\cite{tum} datasets are shown in Figs.~\ref{fig:supp-vis-1},~\ref{fig:supp-vis-2},~\ref{fig:supp-vis-3},~\ref{fig:supp-vis-4} and~\ref{fig:supp-vis-5}. Our method consistently produces accurate inliers across diverse indoor scenes. For example, in the first row of~\ref{fig:supp-vis-3}, existing methods struggle with a high proportion of outliers, while our method achieves an 80.5\% inlier ratio. In Fig.~\ref{fig:supp-vis-5}, our method also demonstrates stable performance across different timestamps. These results further confirm the effectiveness and robustness of the proposed method. We also present a failure case in Fig.~\ref{fig:supp-failure}. Ex-Sim(3)-Reg relies on several hyperparameters, including the sub-intervals $K$, the scale interval, and the reprojection error threshold $\delta_{2d}$. If these hyperparameters are not properly tuned, the method may fail to distinguish outliers with small reprojection errors. Future work will focus on reducing the sensitivity to hyperparameter settings. 

\begin{table}[t]
\setlength{\tabcolsep}{2pt}
\centering
\caption{
Registration recall of the pruning methods with different monocular depth estimation models on the TUM Freiburg\_01 scene~\cite{tum}. 
}
\resizebox{0.8\linewidth}{!}{
\begin{tabular}{c|ccc}
\toprule
Models & {MASt3R}~\cite{mast3r} & {Depth Anything v2}~\cite{dep_anything_v2} & MoGe v2~\cite{moge_v2} \\
\midrule
Sim(3)-TurboReg~\cite{turboreg} & 0.092  & 0.118 & 0.145 \\
Our method               & \textbf{0.264}  & \textbf{0.277} & \textbf{0.323} \\
\bottomrule
\end{tabular}}
\label{table:exp_mde_models_v2}
\end{table}


\begin{figure}[t]
	\centering
		\includegraphics[width=1\linewidth]{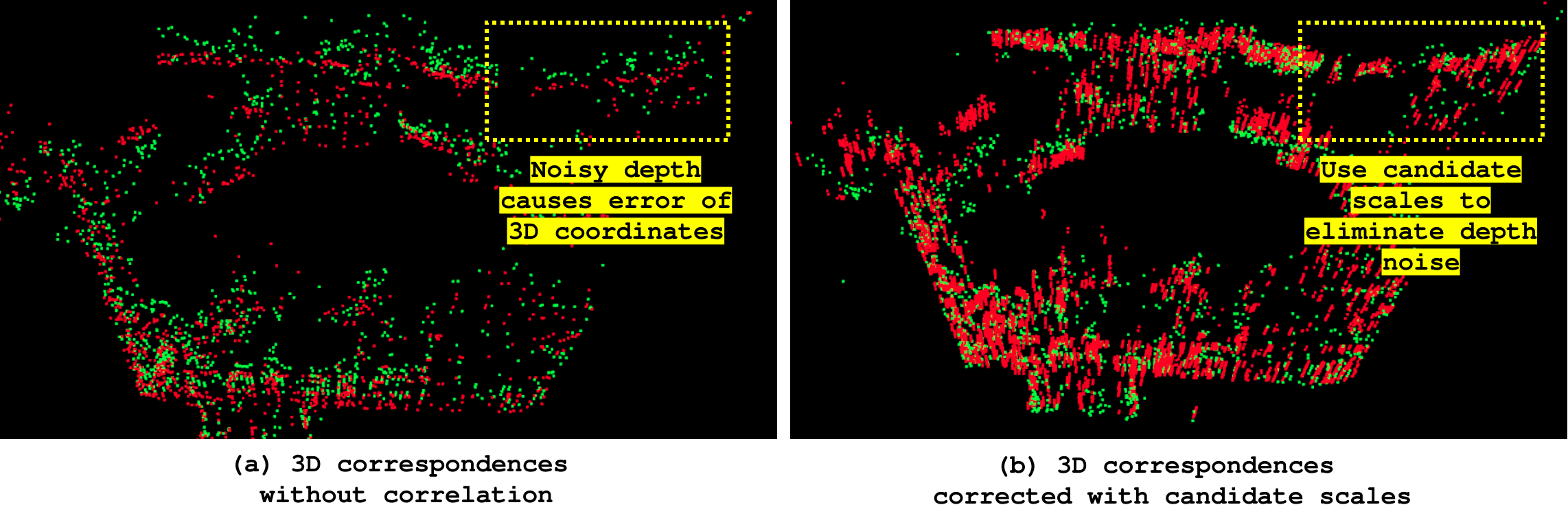}	
        \caption{\textbf{Effect of the proposed scale-aware strategy}.
        (a) When all depth predictions share a single global scale, the transformed 3D points (shown in \textcolor{Red}{red}) do not align well with their corresponding reference points (shown in \textcolor{Green}{green}), revealing noticeable scale inconsistency.
        (b) By generating 3D points with multiple scales and selecting the optimal scale for each correspondence, the red points align better with the green ones. This correspondence-wise adjustment effectively  reduces scale distortion and explains why the proposed extended Sim(3) based pruning method is more robust than existing approaches based on the regular Sim(3) registration.}
	\label{fig:supp-support}
\end{figure}

\section{Verifications of Different MDE Models}
 
To investigate how pruning performance varies with monocular depth estimation (MDE) models, we evaluate the models listed in Table~\ref{table:mde_models}. Experiments are conducted on the TUM Freiburg\_01 scene~\cite{tum}, with results presented in Table~\ref{table:exp_mde_models}. The proposed method remains robust across all MDE models and consistently achieves the highest registration accuracy. Additionally, we analyze pruning performance under varying levels of depth prior accuracy using three MDE methods: MASt3R~\cite{mast3r}, Depth Anything V2~\cite{dep_anything_v2}, and MoGe V2~\cite{moge_v2} (all using their \texttt{vitl} variants). As shown in Table~\ref{table:exp_mde_models_v2}, our method achieves the best pruning performance across all depth estimation models.

\section{Other Explanation of Our Method}

Existing regular Sim(3) registration based pruning methods search for inliers from 3D correspondences $\{(\tilde{s}\cdot\mathbf{q}_i, \mathbf{p}_i)\}_{i=1}^N$. However, they assume that all predicted depths contain little noise. This leads to misalignment of the 3D correspondences $\{(\hat{s}\cdot\mathbf{q}_i, \mathbf{p}_i)\}_{i=1}^N$ even under the ground-truth transformation (see Fig.~\ref{fig:supp-support}(b)). This misalignment renders the maximum clique search unreliable and often results in suboptimal inlier selection.

To alleviate the noise in depth priors, we generate a set of candidate scales $\{\zeta_k\}_{k=1}^K$ and construct multiple scaled versions of the correspondences, such as  $\{(\zeta_1\cdot\mathbf{q}_i, \mathbf{p}_i)\}_{i=1}^N$, ... , $\{(\zeta_K\cdot\mathbf{q}_i, \mathbf{p}_i)\}_{i=1}^N$ (see Fig.~\ref{fig:supp-support}(b)). For each correspondence, we can seek the optimal $\zeta_i^*$ that minimizes
\begin{equation}
\label{eq_exp}
\zeta_i^* = \arg\min_{\zeta_i \in \{\zeta_k\}_{k=1}^K} \Vert \zeta_i \cdot\mathbf{q}_i - (\mathbf{R}\mathbf{p}_i+\mathbf{t}) \Vert_2^2.
\end{equation}
In this way, we effectively compensate for correspondence-wise scale errors and obtain new 3D correspondences $\{(\zeta_i^*\cdot\mathbf{q}_i, \mathbf{p}_i)\}_{i=1}^N$ with reduced scale distortion. In practice, since $\mathbf{R}$ and $\mathbf{t}$ are unknown, direct computation of  $\zeta_i^*$ is not feasible. Therefore, we design the Ex-Sim(3)-Reg algorithm to approximate the procedure in Eq. \ref{eq_exp}. 

\section{Limitations}
 
Although the proposed method outperforms prior methods, it still has limitations. First, the computational cost remains relatively high (the runtime exceeds 70 ms, as shown in Table~\ref{table:runtime}). Future work will explore accelerating the MAC search procedure on multiple extended graphs. Second, the method's performance is sensitive to the choice of hyperparameters (Fig.~\ref{fig:supp-failure} shows that improper hyperparameter settings can cause noticeable degradation). Developing an adaptive hyperparameter tuning mechanism will be an important direction for future research.


%
%
\bibliographystyle{splncs04}
\bibliography{main}
\end{document}